\documentclass[acmsmall, natbib=false]{acmart}
\AtBeginDocument{}
\RequirePackage[maxcitenames=1]{biblatex}
\setcopyright{acmcopyright}
\copyrightyear{2024}
\acmYear{2024}
\acmMonth{12}
\acmJournal{CSUR}
\acmVolume{}
\acmNumber{}
\acmArticle{}
\acmDOI{}

\usepackage{forest}
\usepackage{tikz}
\usepackage{lscape}
\usepackage{enumitem}
\usepackage{indentfirst}
\usepackage{multirow, multicol}
\usepackage{vcell}
\usepackage{amsmath}
\usepackage{algorithm}
\usepackage[noend]{algorithmic}
\usepackage{graphicx}
\usepackage{color}
\usepackage{chngcntr}
\usepackage{tabularx}
\counterwithin{table}{section}
\counterwithin{figure}{section}
\counterwithin{algorithm}{section}

\usepackage{subcaption}
\usepackage{chngcntr}
\usepackage{booktabs}
\usepackage{multirow}
\usepackage{url}
\usepackage{balance}
\usepackage{comment}
\usepackage{mathtools}

\usepackage{color}

\def\sub#1{_{\rm #1}}

\DeclareMathOperator*{\argmax}{arg\,max}
\usepackage[short]{optidef}

\begin{document}

\title{Exploring the Practicality of Federated Learning: 
A Survey Towards the Communication Perspective}
\author{Khiem Le}\authornote{Equal contribution}
\affiliation{
  \institution{Department of Computer Science and Engineering, University of Notre Dame, IN}
  \country{USA}
}
\author{Nhan Luong-Ha}\authornotemark[1]
\affiliation{
  \institution{College of Engineering \& Computer Science, VinUniversity, Hanoi}
  \country{Vietnam}
}
\author{Manh Nguyen-Duc}
\affiliation{
  \institution{Open Distributed Systems, Technische Universität Berlin, Berlin}
  \country{Germany}
}
\author{Danh Le-Phuoc}
\affiliation{
  \institution{Open Distributed Systems, Technische Universität Berlin, Berlin}
  \country{Germany}
}
\author{Cuong Do}
\affiliation{
  \institution{College of Engineering \& Computer Science, VinUniversity, Hanoi}
  \country{Vietnam}
}
\author{Kok-Seng Wong}\authornote{Corresponding author}
\affiliation{
  \institution{College of Engineering \& Computer Science, VinUniversity, Hanoi}
  \country{Vietnam}
}
\renewcommand{\shortauthors}{Khiem et al.}
\authorsaddresses{}


\begin{abstract}
\textbf{Abstract.} Federated Learning (FL) is a promising paradigm that offers significant advancements in privacy-preserving, decentralized machine learning by enabling collaborative training of models across distributed devices without centralizing data. However, the practical deployment of FL systems faces a significant bottleneck: the communication overhead caused by frequently exchanging large model updates between numerous devices and a central server. This communication inefficiency can hinder training speed, model performance, and the overall feasibility of real-world FL applications. In this survey, we investigate various strategies and advancements made in communication-efficient FL, highlighting their impact and potential to overcome the communication challenges inherent in FL systems. Specifically, we define measures for communication efficiency, analyze sources of communication inefficiency in FL systems, and provide a taxonomy and comprehensive review of state-of-the-art communication-efficient FL methods. Additionally, we discuss promising future research directions for enhancing the communication efficiency of FL systems. By addressing the communication bottleneck, FL can be effectively applied and enable scalable and practical deployment across diverse applications that require privacy-preserving, decentralized machine learning, such as IoT, healthcare, or finance.

\end{abstract}

\maketitle

\clearpage
\section{Introduction} \noindent

\begin{figure}[!t]
    \captionsetup{singlelinecheck = false, format= hang, justification=raggedright, font=Large}
    \centering
    \begin{subfigure}[b]{0.45\textwidth}
        \caption{}
        \centering
        \includegraphics[keepaspectratio, width=\textwidth]{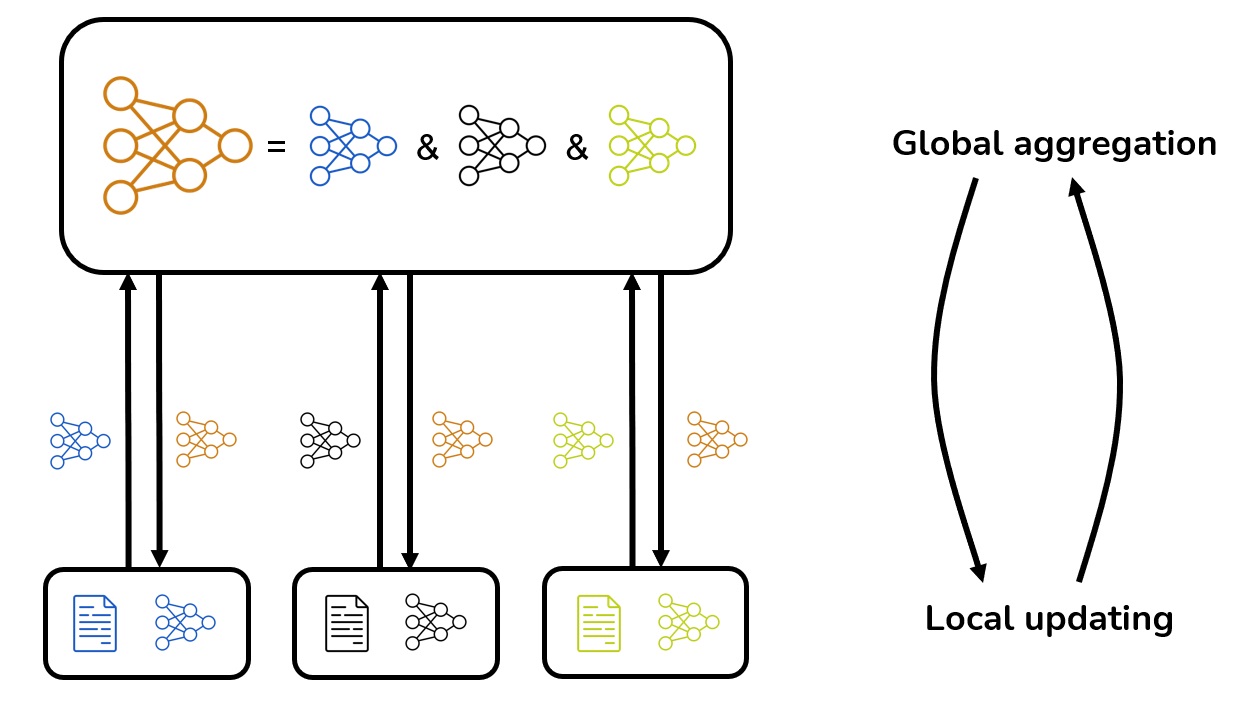}
        \label{fig-fl-workflow}
    \end{subfigure}
    \hfill
    \begin{subfigure}[b]{0.5\textwidth}
        \caption{}
        \centering
        \includegraphics[keepaspectratio, width=\textwidth]{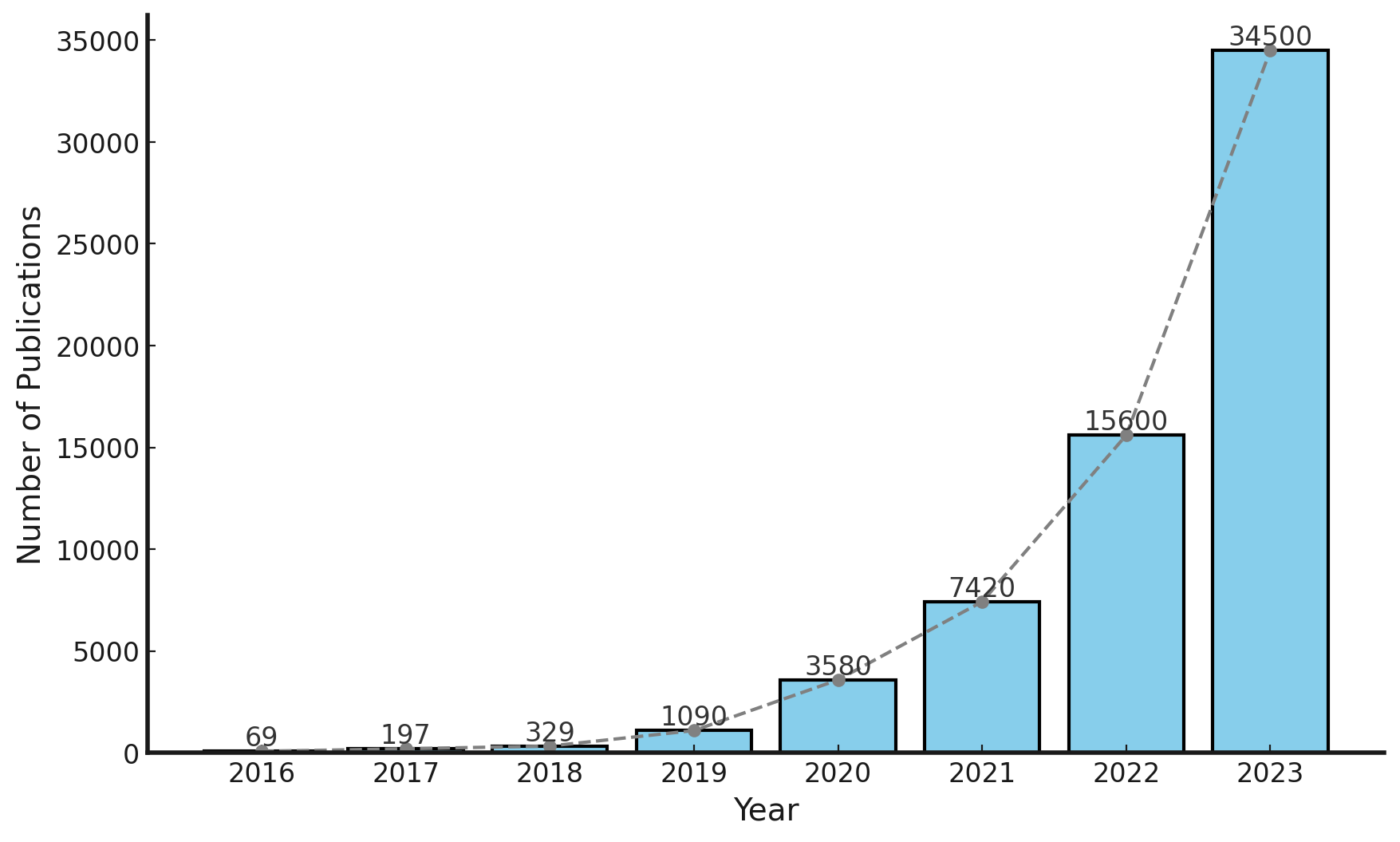}
        \label{fig-fl-publications}
    \end{subfigure}
    \caption{(a) A standard FL system workflow: Client devices train a model based on its private data and upload its trained weights to a central server where all the weights are aggregated and sent back to the clients. (b) Number of articles containing "Federated Learning" \& "Machine Learning" on Google Scholar by year.}
\end{figure}

Federated learning (FL) is a significant development in machine learning, introducing a model focused on data privacy and decentralized processing. Unlike standard machine learning approaches where data is centralized, FL allows for training algorithms across various devices, keeping data localized. This is particularly useful in areas requiring high data privacy, such as healthcare and finance. In FL, data is processed at its source, and only model updates are shared, reducing data transmission and storage risks. This method allows for diverse data sets, capturing real-world variation without centralizing data. The decentralized nature of FL improves privacy and security and supports the development of more personalized and relevant models. Consequently, FL is increasingly used in various applications, including improving mobile device experiences and providing more personalized healthcare, marking a key advancement in secure and efficient data use in machine learning. The recent rise in the popularity and research of FL is evident in the increasing number of publications over the past years. This trend is captured in Fig. \ref{fig-fl-publications}, which illustrates the exponential growth in the number of publications on FL since its inception from 2016 to 2023.\footnote{ Data was sourced from Google Scholar, filtering articles containing both "Federated Learning" and "Machine Learning" from 2016 onward. Articles must contain both terms to be counted towards the tally to avoid articles containing "Federated Learning" from other fields. Citations and patents are excluded. Counts per year are manually recorded based on the reported number of results returned from the query.}

FL faces a significant challenge: communication inefficiency. This arises from the variability in network conditions and device capabilities across many participating devices. Frequent transmission of large model updates across these diverse devices creates a bottleneck, impacting training speed and feasibility. In essence, communication costs overshadow computation costs, hindering the overall efficiency of FL and limiting its practicality in real-world applications \cite{wen2023survey}. The first driving factor behind the focus on communication efficiency is the inherent limitation of network bandwidth, particularly in scenarios involving numerous edge devices such as smartphones and IoT devices. In traditional FL setups, the frequent exchange of model updates between client devices and a central server can quickly become a bottleneck, especially when dealing with large-scale models or operating over limited bandwidth networks. This challenge is further compounded when considering the global expansion of FL applications, where network reliability and speeds can vary significantly. The second factor is the scalability of FL models. As FL finds applications in increasingly complex and data-intensive domains, such as healthcare, autonomous vehicles, and smart cities \cite{yuan2020federated,xu2021federated,singh2022framework,zhang2021end,nguyen2022deep,aivodji2019iotfla,qolomany2020particle,jiang2020federated}, the need for models that can operate effectively and efficiently over large, distributed networks becomes crucial. Scalability in this context is about handling many devices and ensuring that the FL process is robust and efficient regarding communication load and data transfer costs. Lastly, the advancement in algorithms specific to communication efficiency has played a pivotal role. Many techniques have emerged as key solutions to reduce the size of the data transmitted during the FL process, such as regularization \cite{li2020federated}, client selection \cite{FedCS}, quantization \cite{reisizadeh2020fedpaq}, and sparsification \cite{srivastava2014dropout} and many more. These techniques enable the transmission of essential information while minimizing the loss of model accuracy, thus making FL both feasible and practical in bandwidth-constrained environments. 

In summary, the focus on communication-efficient FL responds to the practical limitations and challenges of deploying FL at scale. By addressing these challenges, communication-efficient FL enhances the feasibility of FL across various applications and ensures that the benefits of decentralized learning can be realized in a wide array of real-world scenarios. This survey aims to delve into the different strategies and advancements made in the realm of communication-efficient FL, highlighting their impact and potential in overcoming the communication hurdles inherent in FL systems.

A brief comparison of our survey with related prior works is presented in Table \ref{fig-comparison}. In particular, this paper presents the following contributions:
\begin{itemize}
    \item We summarize the current research on the practicality of FL and the challenges that hinder the widespread deployment of FL in reality with a specific focus on communication overhead.
    \item We conduct an objective review and comparison of existing FL on-device deployment frameworks, mainly focusing on communication management and efficiency.
    \item We discuss sources of communication inefficiency in FL, including communication frequency, communication network limits, client local computation, and number of participating clients.
    \item We provide a comprehensive and up-to-date taxonomy of communication-efficient FL methods built from the system perspective.
    \item We provide valuable insights and lessons from our literature review and empirical results. These findings shed light on promising directions for enhancing the development of high-performance and communication-efficient FL methods.
\end{itemize}

\begin{table}
\caption{Comparison of related surveys. }
\label{fig-comparison}
\resizebox{\textwidth}{!}{%
\begin{tabular}{@{}ccclcccclccc@{}}
\toprule
\multirow{4}{*}{\begin{tabular}[c]{@{}c@{}}Related\\ Surveys\end{tabular}} & \multirow{4}{*}{Focus} & \multicolumn{10}{c}{Communication-efficient FL Methods}                                                                                                                                                                                                                                                                                                                            \\ \cmidrule(l){3-12} 
                                                                           &                        & \multicolumn{8}{c}{Centralized FL}                                                                                                                                                                                                                                                                           & \multicolumn{2}{c}{Decentralized FL}                                \\ \cmidrule(l){3-12} 
                                                                           &                        & \multicolumn{2}{c}{\begin{tabular}[c]{@{}c@{}}Reducing No. of Comm. \\ Rounds\end{tabular}} & \multicolumn{2}{c}{\begin{tabular}[c]{@{}c@{}}Reducing No. of Participating\\ Clients\end{tabular}} & \multicolumn{4}{c}{\begin{tabular}[c]{@{}c@{}}Reduce Network\\ Burdens\end{tabular}}                     & \multirow{2}{*}{Hierarchical FL} & \multirow{2}{*}{Peer-to-Peer FL} \\ \cmidrule(lr){3-10}
                                                                           &                        & Preferring Local Updating              & \multicolumn{1}{c}{One-shot Updating}              & Greedy Selection                                 & Stochastic Selection                             & Quantization         & Sparsification       & \multicolumn{1}{c}{Factorization} & Federated Distillation &                                  &                                  \\ \midrule
\cite{zhao2023towards}                                                     & Communication          &                                        & \multicolumn{1}{c}{}                               &                                                  &                                                  & \checkmark           & \checkmark           & \multicolumn{1}{c}{}              & \checkmark             &                                  &                                  \\ \midrule
\cite{shahid2021communication}                                             & Communication          & \checkmark                             & \multicolumn{1}{c}{\checkmark}                     & \checkmark                                       & \checkmark                                       & \checkmark           & \checkmark           & \multicolumn{1}{c}{}              &                        &                                  & \checkmark                       \\ \midrule
\cite{almanifi2023communication}                                           & Communication          & \multicolumn{1}{l}{}                   &                                                    & \checkmark                                       & \checkmark                                       & \checkmark           & \checkmark           &                                   & \checkmark             & \checkmark                       & \multicolumn{1}{l}{}             \\ \midrule
\cite{mang2023heterogeneous}                                               & Heterogeneity          & \checkmark                  &                                                    & \checkmark                             & \checkmark                             & \multicolumn{1}{l}{} & \checkmark &                                   & \checkmark   & \checkmark             & \checkmark            \\ \midrule
\cite{bellavista2021decentralised}                                         & Decentralized learning & \checkmark                 &                                                    & \checkmark                             & \checkmark                             & \checkmark & \checkmark &                                   & \multicolumn{1}{l}{}   & \checkmark             & \checkmark             \\ \midrule
\cite{pfeiffer2023federated}                                               & Heterogeneity          & \checkmark                  &                                                    & \checkmark                             & \checkmark                            & \checkmark & \checkmark & \multicolumn{1}{c}{\checkmark} & \checkmark   & \checkmark            & \multicolumn{1}{l}{}             \\ \midrule
Ours                                                                       & Communication          & \checkmark                             & \multicolumn{1}{c}{\checkmark}                     & \checkmark                                       & \checkmark                                       & \checkmark           & \checkmark           & \multicolumn{1}{c}{\checkmark}    & \checkmark             & \checkmark                       & \checkmark                       \\ \bottomrule
\end{tabular}%
}
\end{table}

\begin{figure}[!t]
\centering
\includegraphics[keepaspectratio, width=0.8\textwidth]{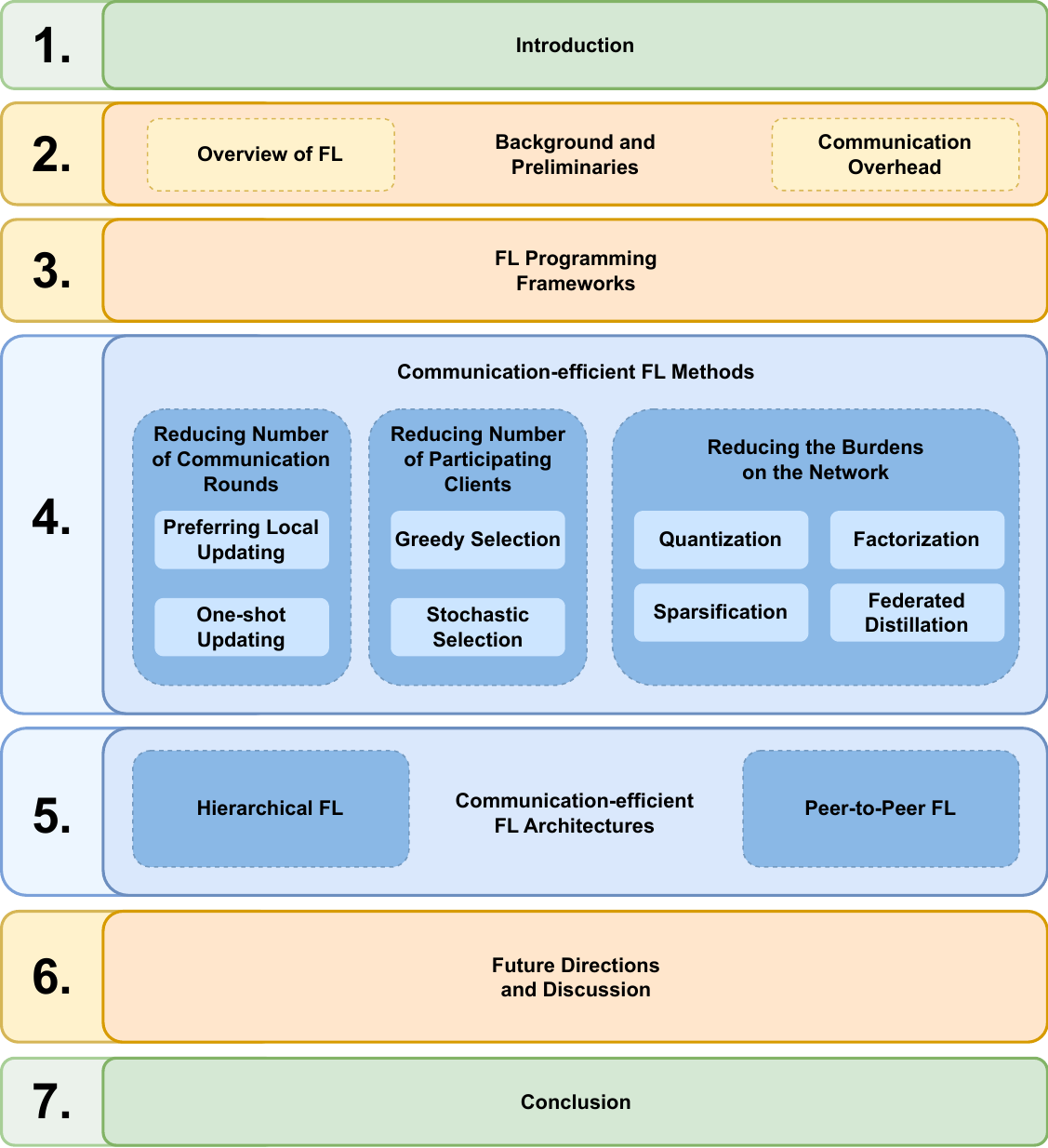}
\caption{Overview of the survey structure. }
\label{fig-structure}
\end{figure}

The structure of this survey is presented in Fig. \ref{fig-structure}. The concept of FL and communication overhead is briefly described in Section 2. Section 3 discusses chosen open-source FL frameworks for experiments. Section 4 surveys the current communication-efficient methods in FL and provides a novel taxonomy. Section 5 presents an overview of decentralized FL architectures. Section 6 provides discussions and future directions. Finally, Section 7 concludes our work. 

\section{Background and Preliminaries} \noindent
To gain a comprehensive understanding of the concepts presented in the survey, this section first provides a motivation for FL, an overview, and workflow of a conventional FL system with two distinct implementations. Then, it highlights the communication process in FL, how the communication overhead can be evaluated quantitatively, and the key factors affecting communication costs in FL systems. 

\subsection{From Centralized Learning to FL}

The fusion of the Internet of Things (IoT) and Artificial Intelligence (AI), often referred to as the Artificial Intelligence of Things (AIoT), has been a significant trend in recent years. This leads to smarter, more connected, and efficient systems. The IoT's network of interconnected devices generates a vast amount of data, which AI processes for intelligent decision-making, automation, and predictive analysis. This combination has transformed various industries, including healthcare, manufacturing, smart cities, and consumer electronics. Initially, most AI systems, particularly those integrated with IoT, relied on \textit{centralized learning} models. In this framework, data collected from various IoT devices were transmitted to a central server where learning algorithms were executed. This approach has several limitations:

\begin{itemize}
    \item \textit{Data Privacy and Security}: Centralized learning requires transferring data from devices to a central server, raising significant concerns about data privacy and security. In sensitive sectors like healthcare, this can be a critical issue.

    \item \textit{Network and Bandwidth Constraints}: Transmitting large volumes of data can be bandwidth-intensive and could lead to network congestion, particularly with the increasing number of IoT devices.

    \item \textit{Latency}: In scenarios where rapid decision-making is crucial (like autonomous vehicles), the delay in sending data to a central server and receiving instructions can be problematic.

    \item \textit{Scalability}: As the number of devices grows, centralized systems may struggle to process the escalating data volumes efficiently.
\end{itemize}

To mitigate some of these issues, \textit{distributed learning} models, where the learning process occurs directly on IoT devices, were introduced. However, these models have their drawbacks:

\begin{itemize}
    \item \textit{Limited Computational Resources}: Many IoT devices have limited processing power, which constrains the complexity and effectiveness of the learning models that can be run on them.

    \item \textit{Inconsistency and Isolation of Learning}: Learning in isolation on individual devices can lead to inconsistent and non-generalized models, as each device only learns from its data.

    \item \textit{Data Diversity and Quantity Issues}: Limited data on individual devices can impede the model's ability to learn effectively, as AI models often require large and diverse datasets.
\end{itemize}

FL emerged as a solution to the challenges posed by both centralized and distributed learning models. Fig. \ref{fig-types} illustrates similarities and differences between the three paradigms. In FL, learning models are trained across multiple decentralized devices or servers holding local data samples without exchanging them. This concept has several advantages:

\begin{itemize}
    \item \textit{Privacy and Security}: Since data remains on the device, FL significantly enhances privacy and security, addressing one of the main drawbacks of centralized learning.
    
    \item \textit{Efficient Use of Bandwidth}: FL reduces the need to transmit large data sets over the network, conserving bandwidth.
    
    \item \textit{Lower Latency}: FL can facilitate quicker decision-making, crucial for real-time applications by processing data locally.
    
    \item \textit{Scalability and Flexibility}: FL can be scaled across numerous devices, making it adaptable to various sectors and applications.
    
    \item \textit{Improved Data Utilization}: FL enables learning from a broader range of data without requiring data centralization, leading to more robust and generalized models.
    
\end{itemize}

FL is a viable solution that harnesses the strengths of both centralized and distributed systems while addressing their key challenges. The introduction of FL marks a significant stride in the AIoT industry, promising enhanced privacy, efficiency, and scalability for AIoT technology. The following subsection will provide an introduction to the fundamentals of FL and its basic implementation.

\begin{figure}[!t]
\centering
\includegraphics[keepaspectratio, width=\textwidth]{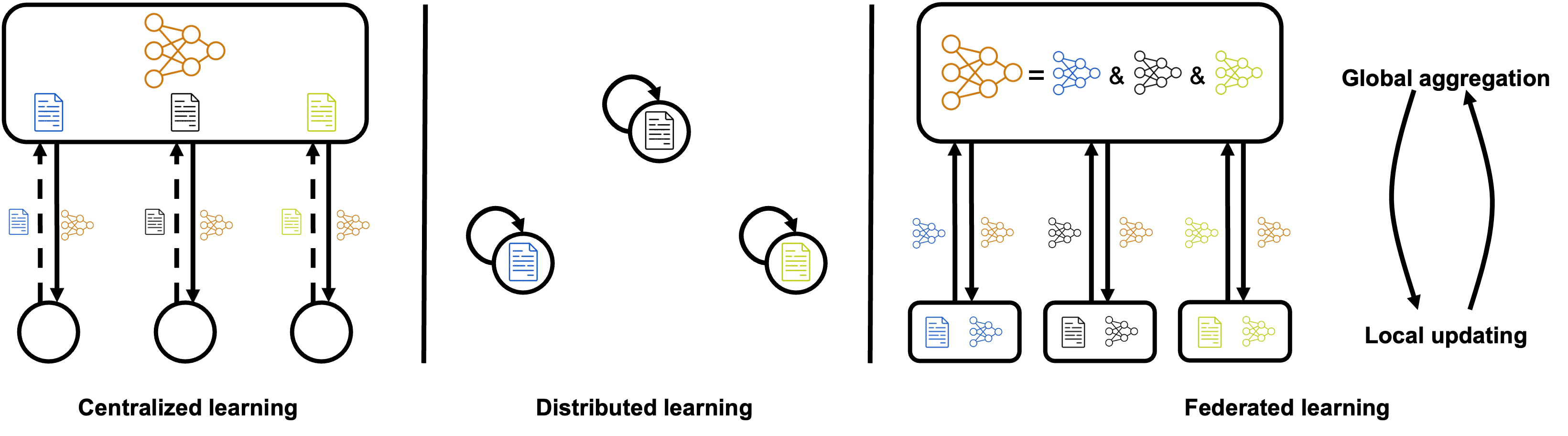}
\caption{Overview of IoT deep learning paradigms. TODO: Update distributed learning section to include model icon}
\label{fig-types}
\end{figure}

\subsection{An Overview of FL}
The key idea of FL is to allow $K$ distributed clients to collaborate and train an ML model without sharing their private data, therefore preserving the data privacy of participating parties. In particular, the considered objective function is formulated as: 
\begin{equation}
\label{eqn-objective}
\underset{w}{\text{min}}\hspace{0.1cm}F(w)=\sum_{k=1}^{K}\frac{n_k}{n} f_k(w),
\end{equation}
where $w$ is parameters of the target ML model, $f_k(w)$ is the empirical risk of $k^{\text{th}}$ client on its private dataset $D_k = \{x_k, y_k\}$, $n_k = |D_k|$ is the number of samples in $D_k$ and $n = \sum_{k=1}^{K} n_k$. To optimize this objective function and achieve a global model representative of the overall data distribution across clients, an FL system utilizes a central server to orchestrate a decentralized training process. At the start, the central server $S$ initiates the global model’s parameters $w^0$ and essential hyper-parameters, such as the number of communication rounds $T$ and the number of participating clients $K$. The server then distributes $w^0$ to all clients and triggers an iterative procedure which generally consists of two main steps: 
\begin{itemize}[leftmargin=*]
\item \textit{Step 1: Local Updating}. Participating clients receive the latest global model’s parameters $w^{t-1}$ from the server $S$, where $1 \leq t \leq T$ is the index of the current communication round. Next, each client $k$ locally trains the model on its private dataset $D_k$. Upon the completion of the local training, client $k$ submits either the gradients $g_k^t = \nabla f_k(w^{t-1})$ or the updated parameters $w_k^t$ to the server to perform aggregation, depending on the implementation. 
\item \textit{Step 2: Global Model Aggregation}. The server $S$ receives either the gradients $g_k^t$ or the updated parameters $w_k^t$ from clients according to the implementation and then fuses the global model to create new parameters $w^t$. Subsequently, $w^t$ is distributed back to clients for conducting the next round of training. 
\end{itemize}

\begin{figure}[!t]
\centering
\includegraphics[keepaspectratio, width=\textwidth]{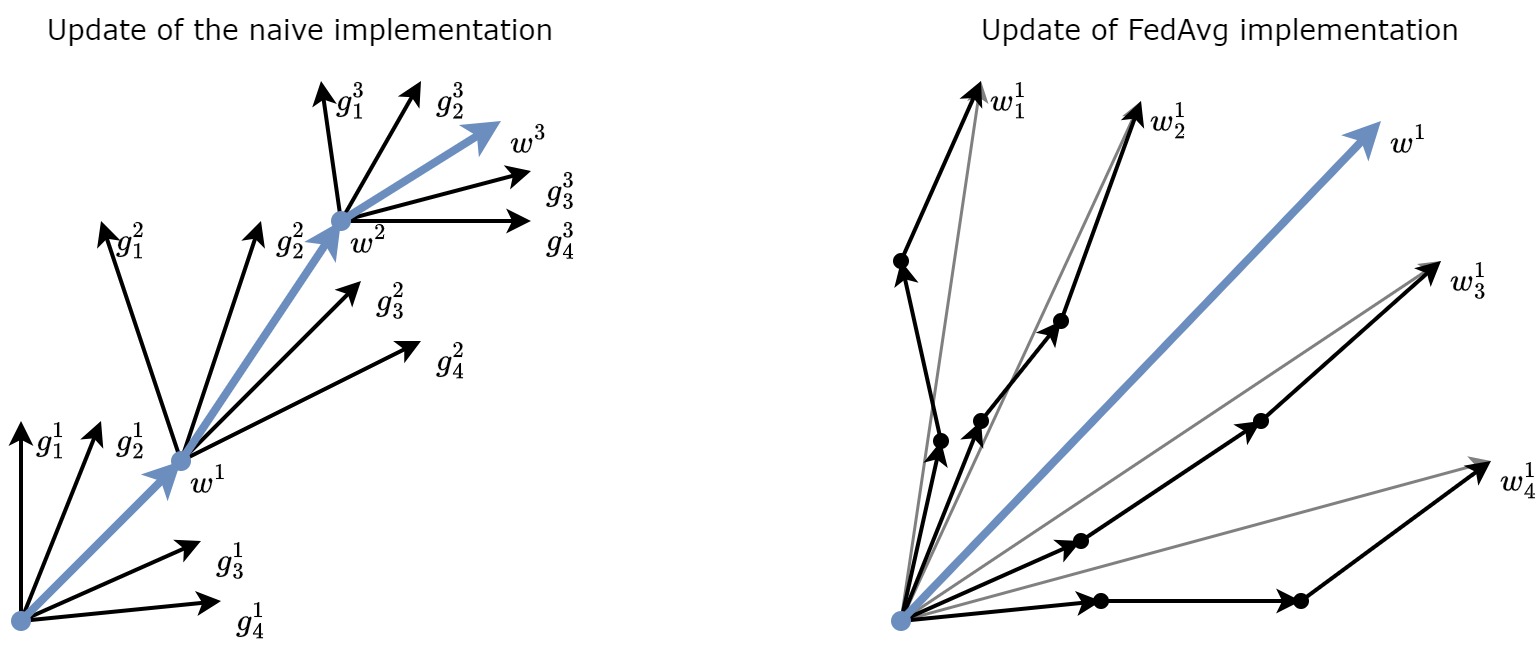}
\caption{An illustration of two distinct implementations in FL containing 4 participating clients and 3 updates. The naive one includes 3 rounds with 1 update each, while the FedAvg employs 1 round with 3 local updates. }
\label{fig-implementations}
\end{figure}

In the naive implementation of FL, each client computes local gradients $g_k^t$ concerning the latest global model’s parameters $w^{t-1}$ and submits $g_k^t$ to the server. The server $S$ aggregates all submitted gradients and then tunes the global model with a predefined client learning rate $\lambda$. The parameters submission and distribution process is repeated over many rounds to make the trained model converge, resulting in expensive communication overhead in the FL system. Different from the naive one, the second one, FedAvg \cite{mcmahan2017communication}, allows each client to perform more than one update on its local data and submits the updated parameters $w_k^t$ rather than the gradients to the server. Next, the server creates the new global model by only aggregating all submitted parameters. Fig. \ref{fig-implementations} shows an illustration to compare these two distinct implementations of FL. Due to the ability to reduce the number of communication rounds and reduce the communication overhead that meets our study scope, we use the second implementation, FedAvg, as the baseline implementation throughout the survey. 

The conventional FL system, as described above, is referred to as a centralized FL or cloud-based FL topology, which involves a central server orchestrating the entire training process. Other advanced decentralized FL topologies will be discussed in later sections. 

\subsection{Communication Overhead and Evaluation}
As described above, unlike traditional ML, which centralizes scattered data to a center and trains a model universally, FL trains the model locally at clients and periodically exchanges the model’s parameters between clients and the server instead of raw data. This parameter exchange process, including parameter submission and distribution, is defined as the communication process in FL and imposes a significant overhead, which is directly proportional to the number of communication rounds. This overhead is even more pronounced when the number of participating clients increases and when there are network conditions constraints, resulting in high power consumption and additional telecommunication expenses \cite{garcia2022prototype, wong2023empirical}. Therefore, evaluating and minimizing the communication overhead in FL is an unavoidable objective. 

To evaluate the communication overhead in FL quantitatively, the total amount of submission and distribution time between clients and the server during the entire FL process, referred to as $cost_{\texttt{time}}$ which is measured in seconds is widely used \cite{asad2021evaluating, cleland2022fedcomm}. Notably, $cost_{\texttt{time}}$ is a metric sensitive to the power of hardware and other implicit factors. A fair environment must be created to use this metric for comparing different systems and methods. In addition to the total time amount, due to the insensitivity, the total size of submitted and distributed parameters, referred to as $cost_{\texttt{size}}$, which is measured in bytes, is also used to evaluate the communication overhead. Formal definitions of these metrics are provided below: 
\begin{equation}
\label{eqn-metrics}
\begin{matrix}
\displaystyle cost_{\texttt{time}}=\sum_{t=1}^{T}\sum_{k=1}^{K}    (\texttt{time}(\rightarrow S,w_k^t)+\texttt{time}(S \rightarrow,w^t)) \\[0.5cm]
\displaystyle cost_{\texttt{size}}=\sum_{t=1}^{T}\sum_{k=1}^{K}    (\texttt{size}(\rightarrow S,w_k^t)+\texttt{size}(S \rightarrow,w^t)) 
\end{matrix}
\end{equation}
where $\texttt{time}(\rightarrow S,\cdot)$ and $\texttt{time}(S \rightarrow,\cdot)$ are the needed time to submit $\boldsymbol{\cdot}$ and distribute $\boldsymbol{\cdot}$ to and from the server $S$, respectively. Also, similar definitions for $\texttt{size}(\rightarrow S,\cdot)$ and $\texttt{size}(S \rightarrow,\cdot)$. Consequently, by minimizing one of these two metrics, the communication overhead can be minimized accordingly.

\subsection{Sources of Communication Inefficiency} \noindent
The communication overhead incurred in FL has been well documented in the existing research literature as one of the core challenges for FL deployment \cite{bellavista2021decentralised, kairouz2021advances}. To propose solutions for overcoming this challenge, the primary sources causing it and their origins need to be identified concretely. By looking at metric definitions in the equation (\ref{eqn-metrics}), we logically derive three major sources of the communication overhead in FL, including the number of communication rounds $T$, the number of participating clients $K$, and the burdens on the network which are represented by $\texttt{size}(\rightarrow S,\cdot)$ and $\texttt{size}(S \rightarrow,\cdot)$.

\begin{table}
\centering
\normalsize
\caption{A summary of communication inefficiency sources in FL. }
\label{tab-sources}
\begin{tabular}{lllcccc@{}}
\toprule
\multicolumn{2}{l}{\multirow{2}{*}{}}                          & \multicolumn{2}{c}{Computing} & \multicolumn{2}{c}{Significance} \\ \cmidrule(l){3-6} 
\multicolumn{2}{l}{}                                           & Hardware      & Algorithm     & Primary         & Secondary      \\ \midrule
\multicolumn{2}{l}{Number of Comm. Rounds}                     &               & \checkmark    & \checkmark      &                \\ \midrule
\multirow{3}{*}{Comm. Network Limits} & Network Bandwidth      & \checkmark    &               & \checkmark      &                \\ \cmidrule(l){2-6} 
                                      & Network Bottleneck     & \checkmark    &               & \checkmark      &                \\ \cmidrule(l){2-6} 
                                      & Transmission Stability & \checkmark    &               & \checkmark      &                \\ \midrule
\multicolumn{2}{l}{Number of Participating Clients per Round}  &               & \checkmark    & \checkmark      &                \\ \midrule
\multirow{2}{*}{Server Computation}   & Computational Power    & \checkmark    &               &                 & \checkmark     \\ \cmidrule(l){2-6} 
                                      & Computational Stress   & \checkmark    &               &                 & \checkmark     \\ \midrule
\multirow{2}{*}{Client Computation}   & Computational Power    & \checkmark    &               &                 & \checkmark     \\ \cmidrule(l){2-6} 
                                      & Computational Stress   & \checkmark    &               &                 & \checkmark     \\ \midrule
\multicolumn{2}{l}{Training Model Scale}                       &               & \checkmark    & \checkmark      &                \\ \bottomrule
\end{tabular}
\end{table}

Table \ref{tab-sources} provides a comprehensive view of the various factors contributing to communication inefficiency in FL.
Number of Communication Rounds $T$ is affected by the FL algorithm chosen and is a primary factor impacting communication costs $cost_{\texttt{time}}$ and $cost_{\texttt{size}}$. Communication network limits are primary factors that hardware affects because the physical network infrastructure is a critical bottleneck that must be considered when designing FL systems. The number of Participating Clients per Round $K$ is similar to $T$, as it is determined by the algorithm chosen and is a primary factor affecting communication costs. Server Computation and Client Computation are considered factors with secondary significance, with computational power and stress as key issues, mainly affected by hardware. Finally, the Training Model Scale is a primary factor, and similar to $T$ and $K$, is determined by the algorithm.

\section{FL Programming Frameworks} \noindent
Communication in FL is cumbersome, especially in real-world settings where clients collaborate in wireless environments. In recent years, thanks to the rapid development of programming tools and frameworks, this process has been managed more efficiently and safely, promoting the feasibility of FL in practice. In this section, we first objectively review representative open-source FL frameworks, mainly concentrating on how they manage the communication process between clients and the server. Then, we further conduct a compact set of experiments to compare their performance practically in an IoT environment. 

Since demonstrating its large potential in 2017 with the introduction of FedAvg \cite{mcmahan2017communication}, FL has become one of the fastest-growing areas of AI during the last five years, proven by the vast number of research publications. To achieve that success, early introduced programming frameworks such as TensorFlow Federated \footnote{https://www.tensorflow.org/federated} and PySyft \cite{ziller2021pysyft} played essential roles. However, these initial frameworks only support simulating FL systems on a single machine, which does not reflect real-world settings. For that reason, more and more advanced frameworks have been introduced in recent years with the support of on-device deployment, which is the most important function for the feasibility of FL in practice. Here, we select a set of available ones to objectively review various aspects, hence providing meaningful suggestions on framework selection for practitioners and researchers. First of all, we only choose frameworks that are released along with a research article describing their architectural design. Therefore, the well-known PaddleFL \footnote{https://github.com/PaddlePaddle/PaddleFL} is excluded due to its lack of transparency in explanation. Next, we use GitRank scores \cite{hasabnis2022gitrank} to rank the remaining ones based on their code quality and maintainability to choose the top five out of seven. Consequently, the OpenFL \cite{reina2021openfl} and FLUTE \cite{garcia2022flute} are excluded from the list because of their lowest scores. Table \ref{tab-frameworks} shows the selected representative frameworks and a detailed qualitative comparison. 

Based on GitRank scores \cite{hasabnis2022gitrank} observed in Table \ref{tab-frameworks}, FedML \cite{he2020fedml} is the leading framework in terms of code quality and maintainability, followed by Flower \cite{beutel2020flower}, which might be attributed to more extended periods of development and maintenance. For usability features, while all selected frameworks provide users with detailed code examples and tutorials, only Flower \cite{beutel2020flower} provides comprehensive API documentation, which is critical for research-purposed customization, whereas FedScale \cite{lai2022fedscale}, NVFlare \cite{roth2022nvidia}, and FederatedScope \cite{xie2023federatedscope} give limited help. All five frameworks support conducting FL systems on IoT devices such as Raspberry Pi and NVIDIA Jetson devices for deployment features. Meanwhile, only FedML \cite{he2020fedml} and Flower \cite{beutel2020flower} support mobile backend and Software Development Kit (SDK) for realistically developing FL on smartphones of both Android and iOS operation systems. In addition, FedScale \cite{lai2022fedscale} claims that its support for iOS-based smartphones is under development and will come soon. 

\begin{table}[!t]
\centering
\tiny
\caption{A qualitative comparison of selected frameworks on usability features and communication backend. An asterisk (*) indicates limited support. This comparison is updated as of December 2023. }
\label{tab-frameworks}

\begin{tabularx}{\textwidth}{@{}lcccccccc@{}}
\toprule
\multirow[t]{2}{*}{Framework}               &
\multirow[t]{2}{*}{Created at}              &
\multirow[t]{2}{*}{GitRank Score}           &
\multirow[t]{2}{*}{Code Example}            &
\multirow[t]{2}{*}{API Documentation}       &
\multicolumn{2}{c}{On-device Deployment}    &
\multirow[c]{2}{*}{\begin{tabular}[c]{@{}c@{}} Communication\\Protocol \end{tabular}} &
\multirow[c]{2}{*}{\begin{tabular}[c]{@{}c@{}} Security\\Protocol      \end{tabular}} \\
\cmidrule(l){6-7}
& & & & & IoT Devices & Mobile Devices & & \\
\midrule
Flower         & 02/2020 & 58.61 & \checkmark \phantom{*} & \checkmark \phantom{*}  & \checkmark \phantom{*} & \checkmark \phantom{*}  & RPC & TLS \\
FedML          & 07/2020 & 65.79 & \checkmark \phantom{*} &                         & \checkmark \phantom{*} & \checkmark \phantom{*}  & MPI &     \\
FedScale       & 04/2021 & 31.48 & \checkmark \phantom{*} & \checkmark*             & \checkmark \phantom{*} & \checkmark*             & RPC &     \\
NVFlare        & 07/2021 & 37.02 & \checkmark \phantom{*} & \checkmark*             & \checkmark \phantom{*} &                         & RPC & TLS \\
FederatedScope & 03/2022 & 40.41 & \checkmark \phantom{*} & \checkmark*             & \checkmark \phantom{*} &                         & RPC &     \\
\bottomrule
\end{tabularx}

\end{table}

From the backend perspective, we observe that gRPC \footnote{https://grpc.io}, a Google-developed version of RPC, is the most common communication protocol adopted by existing frameworks except FedML \cite{he2020fedml}, which employs the MPI protocol. gRPC (Google-developed Remote Procedure Call) is a protocol based on the request-reply model where the client sends a request for the server to execute a specific procedure. It is specially designed for low-bandwidth mobile connections. Conversely, MPI (Message Passing Interface) is based on the message-passing model where processes communicate by explicitly sending and receiving messages and is primarily designed for parallel processing within computers and clusters. According to their communication protocol’s functions, we can conclude that while other frameworks target to support FL across large-scale devices, FedML \cite{he2020fedml} is initially built toward cross-silo FL. Notably, although initially employing the MPI protocol, FedML \cite{he2020fedml} recently supports other various protocols such as MQTT (Message Queue Telemetry Transport) and even gRPC to meet the different demands for the performance of connections, offering users unique flexibility for dealing with various scenarios. In addition to the communication protocol, Flower \cite{beutel2020flower} and NVFlare \cite{roth2022nvidia} are two frameworks that enhance communication security by using the TLS (Transport Layer Security) protocol with CA (Certificate Authority) certificates.

\section{Communication-Efficient FL Methods} 
\begin{figure}[!t]
\centering
\includegraphics[keepaspectratio, width=\textwidth]{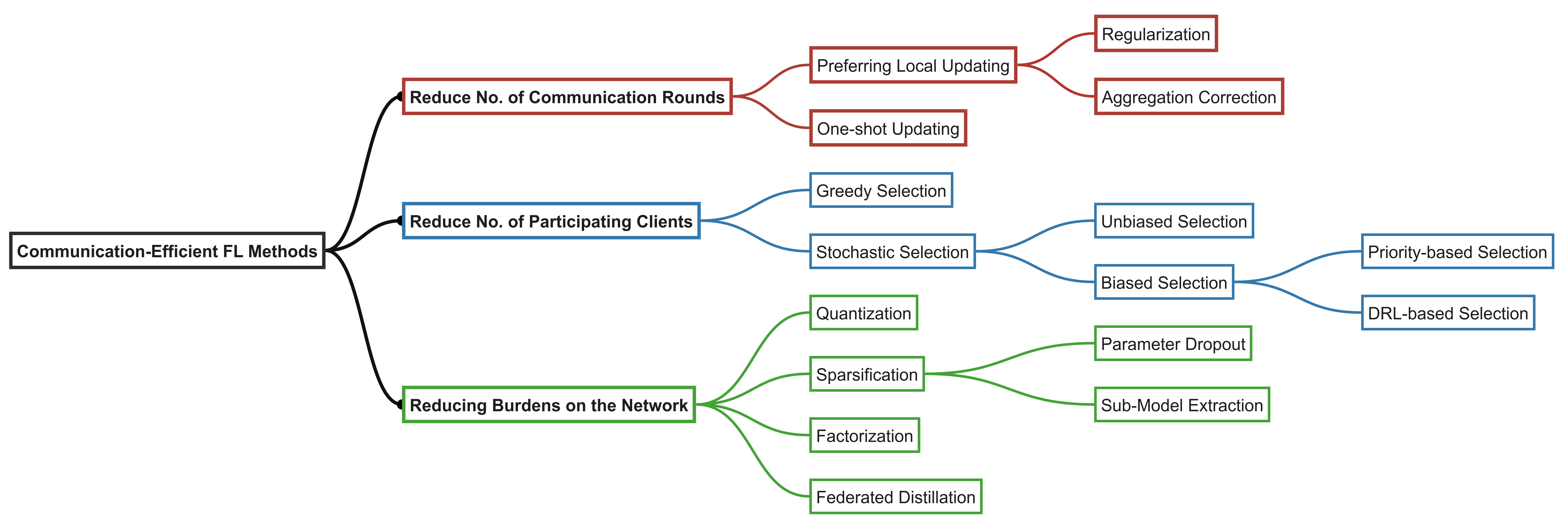}
\caption{A taxonomy of communication-efficient FL methods covered in this survey. Firstly, methods aimed at reducing a number of communication rounds are discussed in Section \ref{sec-number-of-communication-rounds}. Secondly, methods aimed at reducing a number of participating clients (client selection) are discussed in Section \ref{sec-number of clients}. Lastly, methods aimed at reducing burdens on the network via model compression are discussed in Section \ref{sec-burdens on the network}.}
\label{fig-taxonomy}
\end{figure}

\noindent
The research area of improving communication efficiency in FL has been widely explored in recent years with diverse solutions. In this section, we extensively investigate the literature and categorize state-of-the-art methods regarding their objective of tackling discussed sources of inefficiency, including the number of communication rounds $T$, the number of participating clients $K$, and the burdens on the network which are represented by $\texttt{size}(\rightarrow S,\cdot)$ and $\texttt{size}(S \rightarrow,\cdot)$. Within each category, a comprehensive taxonomy of methods is conducted based on the approach used in the solutions rather than their characteristics. Lastly, an empirical benchmark of representative methods in an IoT environment is presented to understand the respective category better. For ease of navigation, Fig. \ref{fig-taxonomy} shows our categorization of this research area. 

\subsection{Reducing the Number of Communication Rounds}
\label{sec-number-of-communication-rounds}
\begin{figure}[!t]
\centering
\includegraphics[keepaspectratio, width=\textwidth]{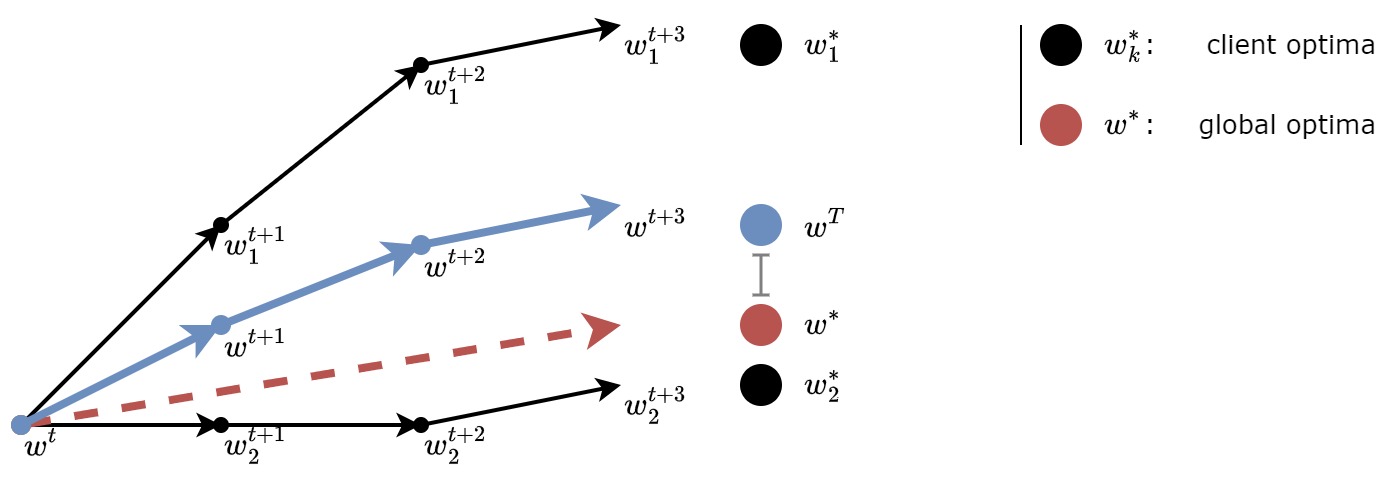}
\caption{An illustration of the client drift issue of FedAvg containing 2 participating clients and 3 local updates. After $T$ communication rounds, the client models move towards their optima $w_k^{*}$ (black circles). In contrast, the global model moves towards $w^T$ instead of to the true optima $w^*$ (red circle), leading to a “\textit{drift}”. }
\label{fig-drift}
\end{figure}

Preferring local updating is a natural solution to reduce the frequency of communication between clients and the server. However, this solution causes the client drift issue, hurting the global model performance, especially under the Non-IID data scenario. As illustrated in Fig. \ref{fig-drift}, client drift refers to the phenomenon where the client models diverge or deviate from the global model over time, leading to a “\textit{drift}” in the optimization of the global model \cite{shi2022optimization}. Thus, methods in this category typically aim to address this associated issue and can be classified into two major groups, i.e., \textit{Regularization} and \textit{Aggregation Correction}. Furthermore, we cover an emerging research direction that targets learning the global model in only one communication round, ideally minimizing the communication overhead. 

\subsubsection{Preferring Local Updating}

\paragraph{Regularization}
The following methods implement variable regularization terms to the local empirical risk, regularizing client models based on different assumptions and observations to overcome the client drift issue. Straightforwardly, \textcite{li2020federated} introduce FedProx, which penalizes client models far from the global model in the parameter space. This is achieved by concurrently optimizing the local empirical risk and the Euclidean distance between the client model’s parameters and the latest global model’s parameters. FedProx has been theoretically and experimentally demonstrated to bring better performance and convergence than FedAvg. Nevertheless, due to the inexactness of minimization, FedProx does not align the local stationary point with the global stationary point. Motivated by that, \textcite{acar2020federated} propose to append an extra term based on exact minimization to ensure that local optima are asymptotically consistent with the stationary point of the global empirical risk. Parallelly, \textcite{charteros2023lay} generalizes FedProx in another way by capturing dissimilarities of the local and global models at a layer basic. Since regularization is applied per layer and not to all parameters as in FedProx, during local training, only the parameters of those layers that drift away from the global model change, while the other parameters are unaffected. Instead of penalizing distances in the parameter space, \textcite{li2021model} constrain the local training of clients by penalizing the distance in the representation space based on an intuitive assumption that the global model trained on a whole dataset can learn a better representation than the local model trained on a skewed subset. Specifically, MOON is introduced to use contrastive loss \cite{khosla2020supervised} to decrease the distance between the representation learned by the local model and the representation learned by the global model and increase the distance between the representation learned by the local model and the representation learned by the previous local model. Share the same assumption, \textcite{kim2022multi} optimize Kullback–Leibler divergence of representations learned by the local and global models at multi-level blocks in the model architecture rather than only the last ones like MOON. Beyond the distance-based point of view, \textcite{lee2022preservation} rely on the forgetting view in FL \cite{shoham2019overcoming} and observe that the knowledge outside of local distribution is prone to be forgotten in local training and is closely related to the client drift issue. Then, FedNTD is introduced, which leverages the knowledge distillation (KD) technique \cite{hinton2015distilling} to preserve the global perspective on local data but only for the not-true classes. Solid evidence shows that not-true logits from local data contain enough knowledge to prevent forgetting while avoiding the collapse in the local empirical risk. Furthermore, \textcite{wu2023fednp} recently presented a unique perspective on handling Non-IID data, which considers the global data distribution when performing local training. Based on that, FedNP is proposed to enhance the local training task with an auxiliary task that explicitly estimates a latent global data distribution, stabilizing the process of local training, whereby a probabilistic neural network is adopted to map such distribution to a global model distribution, consequently regularizing the local model by avoiding it sinking into local data distribution. Although applying the expectation-propagation algorithm \cite{minka2001expectation} to facilitate the auxiliary task, FedNP is designed to be differentiable, leading to an efficient and scalable method. 

\paragraph{Aggregation Correction}
Different from ones in the counterpart group, which regularizes client models during local training, methods in this group aim to correct the global model aggregation step. Firstly, \textcite{wang2020tackling} suggests that the client drift issue is the consequence of inconsistency in the number of local updates among clients due to differences in local data amounts and computational resources. Thus, FedNova is introduced, which normalizes and scales the local model according to its number of local updates before updating the global model to ensure that the resulting global model is not biased. Although it only slightly modifies FedAvg, FedNova has also been theoretically and experimentally demonstrated to improve performance and convergence. From another perspective, \textcite{hsu2019measuring} uses the momentum technique at the server to control the moving progress of the global model, hence preventing client models from diverging fast. This is achieved by first overriding FedAvg in the generalized form, FedOpt \cite{reddi2020adaptive}, as thoroughly formalized in Algorithm \ref{alg-FedOpt}, by treating local updates as “pseudo-gradients” and using a separate server learning rate in global model aggregation. Global momentum is applied to these “pseudo-gradients” like that of well-known optimization algorithms such as SGD. Besides, \textcite{reddi2020adaptive} further employ advanced adaptive optimizers such as AdaGrad and Adam to FedOpt to establish FedAdaGrad and FedAdam, respectively. Furthermore, numerous works \cite{wang2022communication, wu2023faster} have significantly improved the adaptivity of FedAdaGrad and FedAdam in different ways. While all of these methods focus on adaptively adjusting client learning rates, \textcite{jhunjhunwala2022fedexp} present that the server learning rate has a significant impact on the convergence of FedOpt, then derive a time-varying bound \cite{pierra1984decomposition} on the progress made by clients towards the global optima and show how this bound can be used to estimate a good server learning rate at each round. The result is a novel method, FedExP, which adaptively determines the server learning rate at each round based on “pseudo-gradients” in that round. Also built upon FedOpt, \textcite{yu2022tct} recently disclosed that the above methods struggle to work with deep neural networks due to their nonconvexity. Interestingly, the failure of existing methods for deep neural networks is not uniform across the layers when early layers learn useful features. Still, the final layers fail to make use of them. Leverage this finding, TCT is introduced to first use FedOpt to conventionally train a deep model to extract useful features, then compute a convex approximation of the deep model using its empirical Neural Tangent Kernel \cite{jacot2018neural}, and use correction methods to train the final model. TCT can be considered an added layer that makes previous methods significantly faster and more stable to hyper-parameters. 

\begin{algorithm}[!t]
\caption{\colorbox{green}{FedOpt} Algorithm \cite{reddi2020adaptive} \\
$K$ participants are indexed by $k$, $D_k$ is the local dataset at participant $k$, $n_k = |D_k|$ and $n = \sum_{k=1}^{K} n_k$. $w$ is parameters of the target ML model, $T$ is the number of rounds, $E$ is the number of local epochs. $\eta$ is the server learning rate. }
\label{alg-FedOpt}
\begin{algorithmic}[1]
\STATE \textbf{Server executes:}
\begin{ALC@g}
\STATE initiate $w^0$
\FOR{each round $t = 1, 2, 3, \dots, T$}
\FOR{each client $k = 1, 2, 3, \dots, K$ \textbf{in parallel}}
\STATE $\Delta_k^t$ $\leftarrow$ ClientUpdate($k$, $w^{t-1}$)
\ENDFOR
\STATE \colorbox{green}{$\Delta^t=\sum_{k=1}^{K} \frac{n_k}{n} \Delta_k^t$}
\hspace*{\fill} // “pseudo-gradients”
\STATE \colorbox{green}{$w^t = w^{t-1} - \eta \Delta^t$}
\ENDFOR
\STATE return $w^{T}$
    
\end{ALC@g}
\STATE ClientUpdate($k$, $w^{t-1}$):
\FOR{each epoch $e = 1, 2, 3, \dots, E$}
\STATE $w_k^t = w^{t-1} - \lambda \nabla f_k(w^{t-1})$
\STATE $\Delta_k^t = w^{t-1}-w_k^t$
\ENDFOR
\STATE return $\Delta_k^t$
\end{algorithmic}
\end{algorithm}

\subsubsection{One-Shot Updating}
One-shot updating for FL has recently emerged as a promising direction for reducing the communication overhead, which allows the central server to learn the global model in a single communication round. Other motivations behind One-shot FL are that traditional multi-round FL is impractical in some scenarios such as model markets \cite{vartak2016modeldb} and frequent communication poses a high risk of the system being attacked \cite{rodriguez2023survey}. In this track, \textcite{guha2019one} were the pioneers who constructed ensemble learning over client models after one round of completion by simply averaging the predictions of each model. Due to leveraging ensemble learning, this initial method does not require the server to conduct training and can be applied to any ML model. Following that, \textcite{li2020practical} propose a novel consistent voting strategy with the support of an auxiliary public dataset to enhance the ensemble. However, given the principle that \textit{many could be better than all} \cite{zhou2002ensembling}, it may not always be the most effective strategy for the server to select all available client models. Hence, \textcite{wang2023data} focus on the ensemble selection problem and find the best subset of client models to the ensemble for the global model, improving the initial method. Specifically, DeDES is introduced, which ensures the diversity of the ensemble by performing clustering among client models and then selecting a representative element in each cluster based on the local validation score or training data amount. Besides the ensemble learning-based approach, \textcite{kasturi2020fusion} relies on a different one that utilizes the single communication round in another way rather than naive training local models, applying statistical techniques to compress clients’ data into compact information and transfer them to the server. Then, the server uses the compact information to generate a synthetic dataset for training the global model. Moreover, \textcite{zhou2020distilled} adopt the advanced data distillation technique \cite{wang2018dataset} to more accurately compress clients’ data, establishing DOSFL. Although demonstrating promising results, these data-based methods send compressed clients’ data to the server, which causes additional communication overhead and potential privacy leakage. To circumvent this limitation, \textcite{zhang2022dense} presents a novel two-stage method, DENSE, which can synthesize a training dataset for the global model without sharing additional information or relying on any auxiliary dataset. In the first stage, DENSE utilizes the ensemble of client models to train a generator, which generates a synthetic dataset for training in the second stage. DENSE uses the ensemble and the generated dataset in the second stage to train the global model. Recently, \textcite{heinbaugh2022data} made use of conditional Variational Auto-Encoders \cite{sohn2015learning} to enhance the dataset generation process. More specifically, two FedCVAE variants are presented, which use client decoders to either ensemble or compactly aggregate for a global decoder. The global decoder then creates the dataset to train a global classifier. Because FedCVAE only shares client decoders, it seems to be more efficient than other relevant methods. 

\begin{table}[!t]
\centering
\normalsize
\caption{A comparison of state-of-the-art One-shot FL methods. }
\label{tab-one-shot updating}

\begin{tabularx}{\textwidth}{@{}lccccc@{}}
\toprule
\multirow[t]{2}{*}{Method}                  &
\multirow[t]{2}{*}{Year}                    &
\multirow[t]{2}{*}{Data-Free}               & 
\multirow[t]{2}{*}{Model Heterogeneity}     & 
\multicolumn{2}{c}{Server-side Computation} \\
\cmidrule(l){5-6}
& & & & \begin{tabular}[c]{@{}c@{}} Ensembling w/o\\a Global Model \end{tabular} & \begin{tabular}[c]{@{}c@{}} Training\\a Global Model \end{tabular} \\
\midrule
\textcite{guha2019one}           & 2019 & \checkmark & \checkmark & \checkmark &            \\
FedKT \cite{li2020practical}     & 2020 &            & \checkmark & \checkmark &            \\
DeDES \cite{wang2023data}        & 2023 & \checkmark & \checkmark & \checkmark &            \\
\textcite{kasturi2020fusion}     & 2020 &            &            &            & \checkmark \\
DOSFL \cite{zhou2020distilled}   & 2020 &            &            &            & \checkmark \\
DENSE \cite{zhang2022dense}      & 2022 & \checkmark & \checkmark &            & \checkmark \\
FedCVAE \cite{heinbaugh2022data} & 2022 & \checkmark & \checkmark &            & \checkmark \\
\bottomrule
\end{tabularx}

\end{table}

\paragraph{\textbf{Discussion}}
In this subsection, we have discussed state-of-the-art solutions for overcoming the client drift issue when preferring local updating to reduce the number of communication rounds. \textit{Regularization} and \textit{Aggregation Correction} are the two primary approaches that have been well explored in the literature. However, both have their limitations. While \textit{Regularization} methods complexify the local training task, thus creating additional memory footprint and computational bottleneck at clients, \textit{Aggregation Correction} methods require the server to perform more computation rather than only aggregating the global model. Lastly, the emerging direction, One-shot updating, is gaining more attention due to its potential to minimize the communication overhead ideally. Table \ref{tab-one-shot updating} compares characteristics of available One-shot FL methods. Current methods are based on ensemble learning or server-side training, putting a heavy computational load on the server. Especially, the latter poses a high risk of the central server being attacked \cite{gong2022backdoor, nguyen2024backdoor}. These discussed problems open numerous challenges and opportunities for future research in reducing the number of communication rounds. 

\subsection{Reducing the Number of Participating Clients}
\label{sec-number of clients}
Client selection in FL strategizes which clients participate in each training round of an FL model. A comprehensive survey by Fu et al. \cite{ClientSelectionSurvey} delves deeply into this, covering fifteen methods and their approaches to system and data heterogeneity. This section synthesizes methods intersecting those in \cite{ClientSelectionSurvey}, identifying common themes and organizing them into subcategories: Greedy Selection (section \ref{sec-methods-2-1}), Stochastic Selection (section \ref{sec-methods-2-2}), which includes Unbiased Selection (section \ref{sec-methods-2-2-unbiased}) and Biased Selection (section \ref{sec-methods-2-2-biased}).

Client selection fundamentally addresses client data distribution, computational capabilities, and network connectivity heterogeneity. These heterogeneous aspects can lead to slow convergence, skewed model accuracy, and concerns over fairness. By strategically selecting a subset of available clients, client selection enhances model accuracy through diverse data contributions, accelerates training by avoiding slower clients, and improves efficiency by choosing clients with adequate computational and network resources.

\subsubsection{Greedy Selection}
\label{sec-methods-2-1}
Greedy selection methods in FL dynamically select clients based on specific criteria in each training round. These methods prioritize \textit{best} clients according to metrics like data quality, computational speed, or network connectivity. However, this approach can compromise full participation and fairness among clients.

FedCS \cite{FedCS} is an early method focusing on client selection, which selects clients based on their resource availability and expected contribution to the model's update. Before distributing the global model, it requests resource information from clients, such as wireless network bandwidth and compute capability. Clients expected to complete the training and upload updates within a specified deadline are chosen for participation in that round. The goal is to allow the server to aggregate as many client updates as possible within a specified deadline in each round ($T\sub{round}$) to accelerate the model training process. The client selection step in each round involves solving an optimization problem. The goal is to maximize the number of selected clients, $\mathbb{S}$, subject to the constraint of meeting a specified deadline. This optimization can be succinctly formulated as: $\max\limits_{\mathbb{S}} |\mathbb{S}|$, such that $T_{\text{round}} \geq T_{\text{cs}} + T_{\mathbb{S}}^\mathrm{d} + \Theta_{|\mathbb{S}|} + T_{\text{agg}}$. Here, $|\mathbb{S}|$ represents the number of selected clients, $T_{\text{round}}$ is the round deadline, $T_{\text{cs}}$ is the time required for client selection, $T_{\mathbb{S}}^\mathrm{d}$ denotes the time to distribute the model to the selected clients $\mathbb{S}$, $\Theta_i$ estimates the time for client $k_i$ to complete the update and upload, and $T_{\text{agg}}$ is the aggregation time. This approach aims to mitigate the straggler effect (where slower clients delay the training process) by selecting clients that can guarantee faster training and communication times. However, while 
  FedCS can significantly accelerate training, but it may introduce bias by preferentially selecting clients with better resources, potentially ignoring the valuable data held by slower, less resourceful clients. 

FedMCCS \cite{FedMCCS} extends the FedCS protocol by incorporating multiple criteria into the client selection process. The paper proposes a bilevel optimization scheme that efficiently selects and maximizes the number of clients participating in each FL round while considering their heterogeneity and limited communication and computation resources. The approach uses stratified-based sampling to filter the available clients and implements an efficient client selection algorithm based on multiple criteria, including CPU, memory, energy, and time. FedMCCS ensures that only clients with sufficient resources and capabilities are selected to participate in the FL process, reducing the risk of failed training tasks and discarded learning rounds that can affect model accuracy. This holistic approach aims to balance fast convergence, model performance, and the imperative for fairness and broad participation. 

Another approach, DivFL \cite{DivFL}, selects a small, diverse subset of clients for model updates using a greedy selection approach to choose a subset of clients based on the marginal gain of the submodular function, which improves convergence, learning efficiency, and fairness. Specifically, given a set of N clients, the goal is to select a subset $S$ of clients that minimizes an objective equivalent to maximizing a submodular facility location function. The stochastic greedy algorithm used to choose $S$ to maximize this function initializes $S \leftarrow \emptyset$ and iteratively adds one client $k^*$ from a random subset of $V \setminus S$, $V^* = \texttt{rand}(V \setminus S, \text{size}=s)$ that maximizes the increase in the number of selected clients until $|S| = K$ where $K$ is the predefined limit on the number of selected clients to restrict communication cost $
S \leftarrow S \cup k^*, \quad k^* \in \argmax_{k \in V^*} [\bar{G}(S) - \bar{G}(\{k\} \cup S)]$. The empirical results show that DivFL outperforms existing methods regarding convergence speed and fairness while reducing communication costs. The authors also provide a theoretical analysis of DivFL's convergence properties, showing that it achieves a faster convergence rate than FedAvg with random client selection. Overall, DivFL offers a promising approach for improving the performance and fairness of FL.

In comparison, while FedCS and FedMCCS primarily focus on resource-based client selection, potentially leading to biases, DivFL emphasizes diversity in client selection. In general, \textit{Greedy Selection} offers an intuitive and clear-cut way of targeting FL training optimization. However, prioritizing clients with better resources in FedCS and FedMCCS might lead to more efficient use of available resources, albeit at the potential cost of fairness and representation. 

\subsubsection{Stochastic Selection}
\label{sec-methods-2-2}
le \textit{Greedy Selection} targets the optimal batch of clients based on predefined criteria, \textit{Stochastic Selection} introduces randomness into the process, potentially increasing diversity and robustness in the participant pool. These methods are split into \textit{Unbiased} and \textit{Biased} approaches. \textit{Unbiased Selection}, does not prefer any particular client over others, aiming to ensure a fair representation. On the other hand, \textit{Biased Selection}, including Importance-based methods and Deep Reinforcement Learning-based (DRL) techniques, deliberately skews the selection towards clients anticipated to offer the most strategic value based on various criteria. 

\paragraph{Unbiased Selection}
\label{sec-methods-2-2-unbiased}
Clustered Sampling \cite{ClusteredSampling} aims to enhance the representativity and stability of the training process by selecting clients to ensure a more balanced representation of data distributions. This technique groups clients into clusters based on their model similarity or sample size and then selects a representative subset from each cluster for training. Clustered Sampling effectively reduces the variance of clients' aggregation weights with two clustering approaches based on sample size and model similarity. This leads to improved representativity and significantly enhances the stability and quality of convergence, particularly in non-IID and unbalanced scenarios. Clustered sampling reduces variance and achieves faster convergence than standard sampling methods. It also improves the representativity and stability of FL training without imposing additional burdens on the client side and can be seamlessly integrated into standard FL frameworks. To achieve unbiasedness, the clustering algorithms are designed to group clients based on specific criteria, and a representative subset of clients is selected from each cluster for training. This selection process aims to ensure that the chosen subset accurately represents the diversity and distribution of data across all clients, thereby reducing bias in the training process. 

OCS \cite{OCS} is another unbiased selection method, the work aims to reduce communication cost by applying a selective communication approach which minimizes the number of transmitted bits, thereby reducing the overall communication overhead. The algorithm initiates by taking the expected batch size, denoted as $m$, as input. Each participating client $i$ in round $k$ independently computes a parallel local update, denoted as $U^k_i$. Subsequently, clients calculate the norm of their updates, represented as $u^k_i = w_i \lVert U^k_i \rVert$, and communicate these norms to the server. The server then computes optimal probabilities, $p^k_i$, based on the received norms and broadcasts these probabilities to all clients. In the final stage, clients independently decide whether to send their updates $w_iU^k_i/p^k_i$ to the server based on the received probabilities, transmitting updates with a probability of $p^k_i$. This selective communication strategy ensures that only clients with \textit{important} updates contribute to the server's update, significantly reducing communication costs in FL. Notably, OCS aims to minimize the variance of the server update for any specified budget $m$ on the number of participating clients, providing an optimal adaptive client sampling procedure that generalizes theoretical results and allows for larger learning rates compared to baseline uniform client sampling. 

Comparatively, while Clustered Sampling focuses on representativity and stability through balanced data distribution, OCS emphasizes communication efficiency and optimal sampling based on update importance.

\paragraph{Biased Selection}
\label{sec-methods-2-2-biased}

\label{tab-biased-heterogeneity}
\begin{table}
\caption{Comparison of FL Heterogeneity tackled by Biased Client Selection methods}
\begin{tabular}{@{}llcccc@{}}
\toprule
Category & Method & Year & Data heterogeneity & System heterogeneity                       \\ \midrule
\multirow{3}{*}{Priority-based} & FOLB \cite{FOLB} & 2021                         & \checkmark & \checkmark   \\
                                            & AdaFL \cite{AdaFL} & 2021                       & \checkmark &              \\
                                            & Power-of-Choice \cite{PowerOfChoice} & 2022     & \checkmark & \checkmark \\
                                            & FedGS \cite{wang2023fedgs} & 2022     & \checkmark & \checkmark \\
                                            & Fed-CBS \cite{Fed-CBS} & 2023     & \checkmark &
                                          
                                           \\ \midrule
\multirow{2}{*}{DRL-based}        & FAVOR \cite{FAVOR} & 2020                       & \checkmark &              \\
                                           & FedMARL \cite{FedMARL} & 2022                   & \checkmark & \checkmark   \\ \bottomrule
\end{tabular}
\end{table}

Here, we discuss two main subcategories of \textit{Priority-based Selection} methods: \textit{Importance-based} and \textit{Deep Reinforcement Learning (DRL)-based}. Table \ref{tab-biased-heterogeneity} shows an overview of the structure for this subcategory and highlights the types of FL heterogeneity each method aims to tackle. Client availability is broadly categorized into \textit{System heterogeneity}.

\paragraph{Priority-based Selection}
\label{sec-methods-2-2-biased-importance}
We will first discuss the \textit{Priority-based} methods, which randomly select the clients using certain \textit{prioritized criteria}. Based on a load balancing strategy by Mitzenmacher \cite{PowerOf2Choices}, Cho et al. showed that their biased client selection method, Power-of-Choice ($\pi_\text{pow-d}$), which preferentially selects clients with higher local losses, achieve faster convergence compared to traditional unbiased client selection strategies in FL \cite{PowerOfChoice}. $\pi_\text{pow-d}$ consists of three main steps. First, a candidate set of $d$ clients is sampled without replacement based on the fraction of data at each client. Second, the current global model is sent to the clients in the candidate set, and they compute and send back their local loss. Finally, the active client set is constructed by selecting the $m$ clients with the highest local losses from the candidate set, with random tie-breaking. These selected clients participate in the training during the next round. The paper also introduces variations of $\pi_\text{pow-d}$: \textit{Computation-efficient variant} $\pi_\text{cpow-d}$ which only uses a mini-batch of uniformly random chosen samples from the local dataset to approximate the local loss, and \textit{Communication-and-computation-efficient variant} $\pi_\text{rpow-d}$ which have the selected clients for each round sends their accumulated averaged loss over local iterations, and the server selects the clients using the latest received value as a proxy for the local loss. 

FOLB \cite{FOLB} is another method that aims at optimizing for convergence in fewer communication rounds. Nguyen et al. also introduce a general methodology for non-uniform client sampling called FedNu, which estimates a decrease in loss from each client's update and selects important clients accordingly. FOLB is the proposed algorithm based on FedNu, in which each device in a multiset $S^t_1$ of devices chosen uniformly at random has its local update $w^{t+1}_k$ and gradient $\nabla F_k(w^t)$ weighted by their alignment with the global gradient, assessed against an unbiased estimate obtained from $S^t_1$. This ensures prioritization of updates contributing significantly to the global learning direction, while a second unbiased estimate of total correlation among the devices from another chosen uniformly at random multiset  $S^t_2$ normalizes these weights. Notably, the aggregation rule for round $t$ of FOLB is as follows:

$$
w^{t+1} = 
w^t + 
\sum_{k\in S^t_1} 
\frac{\langle \nabla F_k(w^t), \nabla_1 f(w^t)      \rangle}
    {\sum_{k'\in S^t_2} \langle \nabla F_{k'}(w^t), \nabla_2 f(w^t) \rangle}
(w^{t+1}_k - w^t),
$$

where $\nabla_i f(w^t)$ is the averaged gradients from multiset $S^t_i$. In short, FOLB grants each chosen device the autonomy to execute its optimization within a set timeframe, after which it sends the updated parameters and computational resource data back to the central server. This approach allows devices to utilize varying amounts of resources, thus effectively adapting to the devices' varying communication and computational capacities.

An attention-based method was proposed by Chen et al. called AdaFL to tackle Non-IID data with adaptive training procedure \cite{AdaFL}. AdaFL consists of two main components: \textit{Attention-based client selection} and \textit{Dynamic fraction for client selection}. The attention-based client selection mechanism improves fairness in training by increasing the chances of participation by clients with poorer local models. This mechanism uses an attention score vector \( a^{(t)} = [a_1^{(t)}, a_2^{(t)}, \ldots, a_M^{(t)}] \) to adjust client selection probabilities, where \( M \) is the total number of clients. It uses the Euclidean distance between the global and local model parameters, \( d_i^{(t)} = \lVert w^{(t+1)} - w_i^{(t)} \rVert^2 \) to adjust the client selection probability, considering the divergence of local models from the global model. The attention scores are updated using
$ a_i^{(t+1)} = \alpha a_i^{(t)} + (1 - \alpha) \cdot \frac{d_i^{(t)}}{\sum_{k \in S_t} d_k^{(t)}} \cdot a_k^{(t)}$,
where \( \alpha \) is a decay rate, and \( S_t \) represents the selected clients in round \( t \). 
The Dynamic fraction for the client selection method balances the trade-off between performance stability and communication efficiency. Instead of a fixed fraction of clients participating in each training round, AdaFL employs a dynamic fraction that progressively increases, improving model accuracy and reducing communication and computation costs. It dynamically adjusts the fraction \( \gamma(t) \) of clients participating in each training round. It is designed to balance communication efficiency and performance stability, with smaller \( \gamma \) values reducing communication costs but potentially slowing convergence and larger \( \gamma \) values enhancing performance stability at the expense of higher communication costs.  The fraction \( \gamma \) is progressively increased from a starting value \( \gamma(1) \) to an ending value \( \gamma(T) \), where \( T \) is the total number of communication rounds. Notably, AdaFL can easily be integrated into existing FL algorithms to achieve better performance with non-IID data. 

Want et al. proposed the Federated Graph-based Sampling (FedGS) method \cite{wang2023fedgs} to address the challenges of bias and instability in FL under arbitrary client availability. FedGS proposes a solution to simultaneously stabilize the global model update and mitigate long-term bias given arbitrary client availability. The method constructs a Data-Distribution-Dependency Graph (3DG) to model the data correlations among clients, ensuring that the sampled clients' data are kept far apart. This graph helps improve the approximation to the optimal model update. The method involves minimizing the variance of the number of times that clients are sampled while also ensuring stable model updates by introducing a constraint related to the far-distance data distribution of the sampled clients. The primary steps involved in the FedGS method include constructing the 3DG to capture the correlations in clients' local data distributions, solving the optimization problem to select clients with balanced client sampling counts, and utilizing the 3DG to stabilize model updates and mitigate long-term bias under arbitrary client availability.


More recently, Zhang et al. \cite{Fed-CBS} proposed Quadratic Class-Imbalance Degree ($QCID$), a privacy-preserving measure for class imbalance, along with a client sampling scheme called Fed-CBS which biases towards clients with the lowest $QCID$. Selecting a subset of clients $\mathcal{M}$ with the grouped dataset $\mathcal{D}_\mathcal{M}^g$ with lowest $QCID(\mathcal{M})$ is equivalent to finding a $\mathcal{D}_\mathcal{M}^g$ having label distribution with closest to being uniform according to $L_2$ distance. Experimental results showed that Fed-CBS is successful in reducing class imbalance. However, it is essential to note that to ensure privacy-preserving property, Fed-CBS requires the adoption of additional methods such as server-side trusted execution environments (TEEs) or Fully Homomorphic Encryption (FHE), which introduces additional computational costs and may not be available to all clients.

\paragraph{DRL-based Selection}
\label{sec-methods-2-2-biased-DRL}
DRL, for the selection of participating clients, has shown great potential in tackling both data heterogeneity and system heterogeneity in FL.  FAVOR \cite{FAVOR} is a Double Deep-Q Learning (DDQL) method that selects client devices to participate in each round of FL using a mechanism based on deep Q-learning. The framework learns to choose a subset of devices in each communication round to maximize a reward that encourages the increase of validation accuracy and penalizes using more communication rounds.  The DRL agent in FAVOR uses the Double Deep Q-Network (DDQN) algorithm to select the best subset of devices for training. The DRL agents maintain the FL privacy principle by requiring only model weight information, not clients' data. The objective of the DDQN used in FAVOR is to learn the optimal action-value function $Q^*(s,a)$ that maps a state-action pair $(s,a)$ to the expected cumulative reward. The DDQN algorithm is an extension of the Q-learning algorithm that addresses the problem of overestimation of action values in Q-learning. The DDQN algorithm uses two separate neural networks, a target network, and an evaluation network, to estimate the Q-values. The target network is used to generate the target Q-values, while the evaluation network is used to select the best action based on the current state. The target network is updated less frequently than the evaluation network to stabilize the learning process. The objective of the DDQN algorithm is to minimize the mean squared error between the target Q-value and the predicted Q-value for each state-action pair. The target Q-value is calculated using the Bellman equation, which considers the immediate reward and the maximum expected future reward for the next state-action pair. The DDQN algorithm aims to learn the optimal policy that maximizes the expected cumulative reward by selecting the best action at each state. In the context of FAVOR, the DDQN algorithm is used to select the best subset of devices to participate in each communication round to minimize the number of communication rounds and improve the validation accuracy. 

 In addition to using DDQN, multi-agent reinforcement learning (MARL) has also been applied to FL client selection by Zhang et al \cite{FedMARL}. While FAVOR focuses mainly on faster convergence on Non-IID data, FedMarl tackles training efficiency by simultaneously optimizing model performance and communication cost. In FedMarl, each client device is represented by a MARL agent at the central server. These agents decide client participation based on the current state, which includes metrics like accuracy improvement, processing and communication latencies. The reward function in FedMarl reflects changes in test accuracy, processing latency, and communication cost, guiding the MARL agents to make optimal client selection decisions. FedMarl demonstrates that trained MARL agents possess high generalizability across different architectures and datasets without retraining. Additionally, the strategy learned by the agents can be inferred directly from experiments, showing that the agents balance between training accuracy and processing latency, with the number of selected clients varying across different training stages. Furthermore, FedMarl adapts its behavior based on the relative importance of objectives like accuracy, latency, and cost. Adjusting the weights in the reward function can lead to different client selection algorithms, indicating FedMarl's flexibility to suit application-specific needs. 
 
Reinforcement learning is an effective method to replace human intuition in tackling optimization problems. DRL-based methods for client selection are a new and promising area that has successfully reduced communication for FL. However, there are still aspects that need to be considered in future DRL-based methods. For example, the training of the DDQL model relies on a single client, which may impede rapid convergence of the agent, while MARL is more computationally demanding as it uses multiple agents.

\paragraph{\textbf{Discussion}}
In this subsection, we have explored a range of methods for client selection aimed at optimizing the FL training process and reducing the number of communication rounds required. These methods can be broadly categorized into two groups: \textit{Greedy Selection} and \textit{Stochastic Selection}. \textit{Greedy Selection} approaches offer a straightforward heuristic to target training optimization for specific metrics. In contrast, \textit{Stochastic Selection} methods focus on selecting clients more fairly to enhance the robustness of FL systems. While much work has been done in client selection, no consensus exists on a universally preferred method. Each approach discussed has demonstrated improvements over naive random client selection. However, it is essential to note that a comprehensive and comparative study of these methods, applied across different FL settings, is currently lacking. Moreover, issues of fairness and security remain essential considerations in client selection. While system heterogeneity mentioned in the works accounts for variations in performance and latencies among client devices, the varying levels of security and privacy associated with each device have not been thoroughly considered. Neglecting these aspects could leave FL systems vulnerable to client-side attacks. Many of the discussed approaches are server-centric, which can introduce centralized vulnerabilities. As such, it is essential to explore decentralized FL methods that may offer enhanced security and robustness by distributing decision-making and reducing the risk of single points of failure. These will be covered in Section \ref{sec-decentralized}.

\subsection{Reducing the Burdens on the Network}
\label{sec-burdens on the network}
As mentioned earlier, using a small ML model rather than a large-scale one is a natural solution to reduce the burdens on the network, which are represented by $\texttt{size}(\rightarrow S,\cdot)$ and $\texttt{size}(S \rightarrow,\cdot)$. However, small models with limited capability struggle to handle complicated tasks and typically perform poorly. Therefore, model compression is a better solution, and the key idea is to shrink large-scale models into smaller sizes while simultaneously maintaining their capability. In this subsection, we investigate the literature on model compression in FL and sequentially discuss the following four major method groups, i.e., \textit{Quantization}, \textit{Sparsification}, \textit{Factorization}, and \textit{Federated Distillation}. 

\begin{algorithm}[!t]
\caption{\colorbox{pink}{FedPAQ} Algorithm \cite{reddi2020adaptive} \\
$K$ participants are indexed by $k$, $D_k$ is the local dataset at participant $k$, $n_k = |D_k|$ and $n = \sum_{k=1}^{K} n_k$. $w$ is parameters of the target ML model, $T$ is the number of rounds, $E$ is the number of local epochs. $\eta$ is the server learning rate. \colorbox{pink}{Q} is the quantizer with the quantization level $s$. }
\label{alg-FedPAQ}
\begin{algorithmic}[1]
\STATE \textbf{Server executes:}
\begin{ALC@g}
\STATE initiate $w^0$
\FOR{each round $t = 1, 2, 3, \dots, T$}
\FOR{each client $k = 1, 2, 3, \dots, K$ \textbf{in parallel}}
\STATE $\Delta_k^t$ $\leftarrow$ ClientUpdate($k$, $w^{t-1}$)
\ENDFOR
\STATE \colorbox{green}{$\Delta^t=\sum_{k=1}^{K} \frac{n_k}{n} \Delta_k^t$}
\hspace*{\fill} // “pseudo-gradients”
\STATE \colorbox{green}{$w^t = w^{t-1} - \eta \Delta^t$}
\ENDFOR
\STATE return $w^{T}$
    
\end{ALC@g}
\STATE ClientUpdate($k$, $w^{t-1}$):
\FOR{each epoch $e = 1, 2, 3, \dots, E$}
\STATE $w_k^t = w^{t-1} - \lambda \nabla f_k(w^{t-1})$
\STATE $\Delta_k^t = \colorbox{pink}{Q}(w^{t-1}-w_k^t, s)$
\ENDFOR
\STATE return $\Delta_k^t$
\end{algorithmic}
\end{algorithm}

\subsubsection{Quantization}
Quantization is the approach that decreases the model size by representing the bit width from a floating point of 32 bits, e.g., \texttt{fp32} to a lower precision, e.g., \texttt{fp16}, \texttt{int32}, \texttt{int16}, \texttt{int8} meanwhile retaining the model performance. In general, quantization methods in the FL scenario employ quantizers on the exchanged model’s parameters to reduce the representing bit, thus reducing the communication overhead in the FL system. At first, \textcite{reisizadeh2020fedpaq} modify FedOpt by applying a stochastic quantizer called QSGD \cite{alistarh2017qsgd} on the submitted “pseudo-gradients” from clients to establish FedPAQ, as shown in Algorithm \ref{alg-FedPAQ}. Straightforwardly, FedPAQ can be viewed as FedOpt with a quantization scheme. Moreover, depending on the quantization level $s$ of QSGD, FedPAQ suffers from a trade-off between reducing the communication overhead and sacrificing the model performance. Following that, \textcite{haddadpour2021federated} enhance the theoretical guarantee of FedPAQ with a tuned server learning rate. Unlike FedPAQ, which has a fixed quantization level $s$ across clients, \textcite{chen2021dynamic} considers diverse computational resources and supports different client quantization levels. Then, FedHQ is presented to assign different aggregation weights to clients by minimizing the convergence upper bound. Furthermore, \textcite{jhunjhunwala2021adaptive} propose an adaptive method, AdaQuantFL, which considers the trade-off between error and communication bits to allow clients to adjust the quantization level during the FL process. More specifically, AdaQuantFL discretizes the entire training process into uniform communication intervals, wherein at each interval, an optimal $s$ is determined so that the training error upper bound is minimized. We can observe that all the above methods quantize the submitted information from clients and assume perfect distribution of the global model from the server. However, it is evident that the total communication overhead can be further decreased if the global model is also encoded before distributing to clients. Based on this motivation, \textcite{amiri2020federated} introduces the Lossy FL, in which QSGD quantizes both the local information and global information before being exchanged. Although stochastic quantizers like QSGD are efficient and convenient for assigning each vector coordinate to a finite set of possibilities, they are sensitive to the vector distribution and the gap between the largest and smallest entries in the vector. Therefore, many studies have concentrated on using other quantizers. For instance, \textcite{suresh2017distributed} devise a deterministic quantization method by applying a structured random rotation before quantization. That is clients and the server draw rotation matrices according to some known distribution, clients then send the quantization of the rotated vectors while the server applies the inverse rotation on the estimated rotated vector. Built upon this method, \textcite{vargaftik2021drive} introduces DRIVE, which could quantize the original vector into a 1-bit quantization level by random rotation. In addition, \textcite{shlezinger2020uveqfed} propose a scheme following concepts from universal quantization \cite{zamir1992universal}, referred to as universal vector quantization for FL, UVeQFed. Specifically, UVeQFed implements subtractive dithered lattice quantization based on solid information-theoretic arguments. Such schemes are known to approach the most accurate achievable finite-bit representation dictated by rate-distortion theory within a controllable gap, as well as achieve more accurate quantized representation compared to stochastic quantization methods used in previous works. 

\subsubsection{Sparsification}
Sparsification is the approach that decreases the model size by generating and exchanging small parts of the original large-scale model between clients and the server at each round. In addition to reducing the communication overhead by generating suitable small parts, this approach is beneficial in assigning resource-adaptive computation to clients. In this track, \textit{Parameter Dropout} and \textit{Sub-Model Extraction} are two major types of methods that generate unstructured and structured client models, respectively. 

\paragraph{Parameter Dropout}
The following methods motivate the classic Dropout technique \cite{srivastava2014dropout} in traditional ML, which randomly drops neurons together with their connections with a drop probability $p$ from the model during training to generate unstructured smaller models for clients. \textcite{wen2021federated} first introduce Federated Dropout, which can be viewed as the “federated version” of Dropout. In particular, the server initially uses the resource budget information from clients to determine a proper drop probability for each one, then applies Dropout with corresponding drop probabilities $\{p_k\}_{k=1}^K$ and assigns the generated models $\{\overline{w_k}\}_{k=1}^K$ to clients for performing local training. Since the generated models $\{\overline{w_k}\}_{k=1}^K$ are sparse after applying Dropout, the communication overhead can be saved significantly for both the submission and distribution processes. To aggregate the global model, the server first establishes $w_k$ where the parameters included in $\overline{w_k}$ are updated, and the other parameters use the same values in the last round. Then, the global model is updated by averaging all complete $w_k$, say $\{w_k\}_{k=1}^K$. Instead of randomly dropping a fraction of neurons from the model, \textcite{bouacida2021adaptive} propose and study Adaptive Federated Dropout, which maintains an activation score map based on the training error to determine which activations should be selected to be transferred or dropped. Each score map assigns real values to all the activations representing their importance and influence on the training process. Moreover, \textcite{liao2021feddrop} incorporate additional layers, namely SyncDrop layers, into the model to determine the channel-wise trajectories to be retained or dropped to automatically adapt the model according to clients’ local data distributions. Parallelly, \textcite{chen2021communication} figure out that many parameters become stable long before the model converges, then after these parameters reach their optimal values, the subsequent updates for them are simply oscillations with no substantial changes and can indeed be safely excluded without harming the model performance. Based on that, another adaptive method is presented, which adaptively freezes and unfreezes parameters to reduce communication volume while preserving convergence. Under this method, each stable parameter after identified is frozen for a certain number of rounds and then unfrozen to check whether it needs further training or not, the length of such freezing period is additively increased or multiplicatively decreased based on whether that parameter is still stable after resuming being updated. Despite reducing the communication overhead, these methods directly manipulate individual model parameters, which negatively affect the scaling efforts and the model performance \cite{cheng2022does}. Motivated by that, \textcite{chen2022fedobd} divide the model into semantic blocks, evaluate block importance rather than determining individual parameter importance, and opportunistically discards unimportant blocks to enable a more significant reduction of communication overhead while preserving the model performance. The resulting method, FedOBD, considers the basic building blocks of deep neural networks as semantic blocks and uses a novel block difference metric to determine block importance. Finally, the global model is aggregated similarly to Federated Dropout. Due to its block-level approach, FedOBD is more efficient for training large-scale models. 

\paragraph{Sub-Model Extraction}
Different from the ones in the counterpart group, which leverages dropout-like techniques to generate unstructured smaller models from the original full model, methods in this group extract and assign structured sub-models for clients, preserving the architectural advances of the entire model. Firstly, \textcite{diao2020heterofl} and \textcite{horvath2021fjord} were the pioneers with their cornerstone methods, HeteroFL and FjORD, respectively. In detail, HeteroFL and FjORD create sub-models by selecting a certain fixed number of kernels of the global model based on clients’ capacities. This means client models are slimmer and different in their layers’ widths but still structured in their architectures. During global model aggregation, the server aggregates kernels using a mechanism similar to Federated Dropout and recovers the full model successfully. Remarkably, under the extraction scheme above, depending on their resource demands, different sub-models can only be trained on clients whose on-device resources are matched. Consequently, part of the global model cannot be trained on data at low-end clients, causing its different parts to be trained on data with different distributions, referring to the undertrained global model issue. This would degrade the model performance, especially under the Non-IID data scenario. Target on resolve this issue, \textcite{hong2021efficient} propose Split-Mix to increase accessible training data by splitting the global model into universally budget-compatible sub-models and re-mix afterward. Under Split-Mix, a client trains all affordable sub-models, which are later ensembled on-demand according to inference requirements. Moreover, \textcite{alam2022fedrolex} propose FedRolex, which uses a creative rolling extraction scheme to address the above issue more efficiently. Specifically, sub-models are extracted from the global model using a rolling window that advances each communication round. Since the window is rolling, sub-models from different parts of the global model are extracted in sequence in different rounds. As a result, all the global model parameters are evenly trained based on the clients' local data. Although demonstrating promising efficiency, width-based splitting schemes tend to result in very slim client models, which can significantly lose basic features, resulting in a drastic drop in model quality \cite{tan2019efficientnet}. Furthermore, these schemes suffer from parameter mismatches of channels in aggregation, leading to a lower performance than simply excluding weak clients from training. From these insights, \textcite{kim2022depthfl} introduces a method based on depth splitting, DepthFL, to solve these issues. In particular, DepthFL defines client models of different depths by pruning the global model's deepest layers and allocating them to clients depending on their resources. Notably, DepthFL alleviates the issue of undertrained deep layers by mutual self-distillation of knowledge among the classifiers of various depths within each client model. Recently, \textcite{ilhan2023scalefl} adaptively scaled down the global model along both width and depth dimensions by leveraging early exits to find the best-fit models for resource-aware local training on clients. The downscaling procedure in the novel method, ScaleFL, is inspired by EfficientNet \cite{tan2019efficientnet}, which demonstrates the importance of balancing the size of different dimensions while scaling deep neural networks. Next, ScaleFL performs self-distillation among early exit and final predictions during local model training to improve the knowledge transfer among sub-models and provide effective aggregation. 

\begin{table}[!t]
\centering
\normalsize
\caption{A comparison of state-of-the-art Sparsification methods. }
\label{tab-sparsification}

\begin{tabularx}{\textwidth}{@{}lccccc@{}}
\toprule
\multirow[t]{3}{*}{Method} & \multirow[t]{3}{*}{Year} & \multicolumn{3}{c}{Client Models} & Model Undertraining \\
\cmidrule(l){3-5}
 & & \multirow[t]{2}{*}{Unstructured} & \multicolumn{2}{c}{Structured} \\
\cmidrule(l){4-5}
 & & & Width & Depth \\
\midrule

Federated Dropout \cite{wen2021federated}              &2021& \checkmark &            &            & \checkmark \\
Adaptive Federated Dropout \cite{bouacida2021adaptive} &2021& \checkmark &            &            & \checkmark \\
\textcite{liao2021feddrop}                             &2021& \checkmark &            &            & \checkmark \\
\textcite{chen2021communication}                       &2021& \checkmark &            &            & \checkmark \\
FedOBD \cite{chen2022fedobd}                           &2022& \checkmark &            &            & \checkmark \\
HeteroFL \cite{diao2020heterofl}                       &2020&            & \checkmark &            & \checkmark \\
FjORD \cite{horvath2021fjord}                          &2021&            & \checkmark &            & \checkmark \\
Split-Mix \cite{hong2021efficient}                     &2021&            & \checkmark &            &            \\
FedRolex \cite{alam2022fedrolex}                       &2022&            & \checkmark &            &            \\
DepthFL \cite{kim2022depthfl}                          &2022&            &            & \checkmark &            \\
ScaleFL \cite{ilhan2023scalefl}                        &2023&            & \checkmark & \checkmark &            \\

\bottomrule
\end{tabularx}

\end{table}

\subsubsection{Factorization}
Factorization is the approach that decreases the model size by decomposing the high-dimensional model’s parameters into low-rank matrices, where such matrices are ensured to be sufficiently informative and approximate the original model. To open, \textcite{konevcny2016federated} enforce the model’s parameters $w_k^t \in \mathbb{R}^{I \times O}$ to be a low-rank matrix of rank at most $r$, where $r$ is a fixed number \footnote{For the sake of simplicity, we suppose that $w_k^t$ is a 2D matrix, for instance, the parameters of a fully connected layer. Moreover, we only discuss the case of a single matrix since everything carries over to a setting with multiple matrices, for instance, corresponding to individual layers in deep neural networks. }. In order to do so, $w_k^t$ is naively expressed as the product of two matrices $w_k^t = w_{k, 1}^t \times w_{k, 2}^t$, where $w_{k, 1}^t \in \mathbb{R}^{I \times r}$ and $w_{k, 2}^t \in \mathbb{R}^{r \times O}$. In subsequent computation, $w_{k, 1}^t$ is randomly generated in each round for each client and considered a constant during local training. In contrast, only $w_{k, 2}^t$ is trained on local data and then exchanged. In this way, this initial method immediately saves a factor of $I/r$ in the communication overhead. However, because of directly training the low-rank $w_{k, 2}^t$ from scratch, the model performance drops significantly when the factorization rate $h$ increases. After that, \textcite{hyeon2021fedpara} introduced a new low-rank Hadamard product factorization called FedPara, which is not restricted to low-rank constraints, and thereby has a far larger capacity. The key idea is combining the Hadamard product with low-rank factorization as $w_k^t = w_{k, 1}^t \times w_{k, 2}^t = (u_{k, 1}^t \times u_{k, 2}^t) \phantom{.} {\scriptstyle \bigodot} \phantom{.} (v_{k, 1}^t \times v_{k, 2}^t)$, where ${\scriptstyle \bigodot}$ represents a Hadamard product. When $\text{rank}(u_{k, 1}^t \times u_{k, 2}^t) = \text{rank}(v_{k, 1}^t \times v_{k, 2}^t) = r$, then $\text{rank}(w_k^t) \leq r^2$. This property facilitates the constructed matrix spanning the full-rank matrix with fewer parameters than the naive factorization. Contrasting with the previous one, FedPara can achieve a performance comparable to that of its original counterpart. Nevertheless, FedPara conducts multiplications many times during training, including the Hadamard product, and these multiplications may potentially be more susceptible to numerical instability. To further reduce the communication overhead, \textcite{jeong2022factorized} utilize rank-1 matrices to communicate and perform aggregation in the lowest subspace possible for compatibility while effectively enhancing expressiveness using local sparse bias matrices. Specifically, the resulting method, Factorized-FL, expresses the model’s parameters as $w_k^t = w_{k, 1}^t \times w_{k, 2}^t \times \mu_k^t$, where $w_{k, 1}^t \in \mathbb{R}^{I \times 1}$ and $w_{k, 2}^t \in \mathbb{R}^{1 \times O}$, and $\mu_k^t \in \mathbb{R}^{I \times O}$ is a highly sparse bias matrix to capture further the information not captured by the outer product of the two matrices. Notably, $\mu_k^t$ is initialized with zeros, and its sparsity is controlled by the hyper-parameter for the sparsity regularizer, so it can gradually capture the additional expressiveness during training. Another crucial difference between Factorized-FL and the previous methods is that it considers the Non-IID data scenario in the aggregation of factorized matrices. In addition to training the low-rank matrices from scratch, \textcite{qiao2021communication} present FedDLR, which performs low-rank factorization once clients finish their local training and then upload the respective low-rank matrices to the server. The server then aggregates the received matrices into a global model and performs another factorization of the aggregated model for distribution to clients. In particular, FedDLR explicitly factorizes the model’s parameters via the truncated Singular Value Decomposition \cite{sadek2012svd}, a common post-training compression technique. By employing dual-side factorization, the communication overhead under FedDLR is monotonically decreasing during the entire training process. However, this puts additional computation workloads on both clients and the server. 

\subsubsection{Federated Distillation}
Federated Distillation is the approach that decreases the model size by leveraging the knowledge distillation (KD) technique \cite{hinton2015distilling} to transfer the acquired knowledge of a large-scale model, which is embedded in the model’s outputs, to a small model without adversely impacting the final performance. First off, \textcite{li2019fedmd} develop FedMD in which a labeled public dataset $D^p = \{x^p, y^p\}$ is introduced to all clients in addition to their private dataset $D_k = \{x_k, y_k\}$. With consecutive training on $D_k$ and inference on $D^p$, clients transfer the knowledge of their private dataset $D_k$ into the predictions $\widehat{y^p}_k$ on $D^p$. The predictions $\widehat{y^p}_k$ are then uploaded to the server and aggregated to result in new labels of $D^p$ for the next round of training. The major difference between FedMD and the baseline FedAvg is that knowledge is not shared through the model’s parameters but through the predictions on a labeled public dataset. In this way, FedMD undoubtedly saves the communication overhead. However, consecutive training on $D_k$ and inference on $D^p$ necessitates considerable local computation and is unsuitable for resource-limited clients. Furthermore, creating a labeled public dataset necessitates careful deliberation and thus lacks generalization. Consequently, \textcite{lin2020ensemble} propose FedDF to move KD from clients to the server, training the global model through an unlabeled public dataset on the outputs of client models in an ensemble distillation manner. FedDF removes additional inference effort from clients and is robust to the public dataset selection. Following that, \textcite{sattler2021fedaux} propose FedAUX, an extension to FedDF, which, under the same set of assumptions, drastically improves the performance by deriving maximum utility from the unlabeled auxiliary dataset. In particular, FedAUX calculates the certainty of each client model’s outputs by contrastive logistic scoring and uses this certainty to determine the aggregation weights to yield a more robust empirical performance on data with high distinction. Similarly, \textcite{li2021personalized} construct an adaptive aggregation step name pFedSD, which dynamically modifies the aggregation weight of each client by exploiting the Jensen-Shannon divergence between the exchanged model’s outputs in two adjacent rounds. Although demonstrating promising results, these data-based methods are not always practical due to a reliance on a labeled or unlabeled public dataset, which might not be available in extreme cases. Thus, \textcite{zhu2021data} introduces a data-free method, where the server learns a lightweight generator to ensemble client knowledge, which is then distributed to clients, regulating local training using the learned knowledge as an inductive bias. Precisely, the introduced method, FedGEN, learns a generative model derived solely from the prediction rules of client models, which, given a target label, can yield feature representations that are consistent with the ensemble of predictions. This generator is later distributed to clients, escorting their local training with augmented samples over the latent space, which embodies the distilled knowledge from other peer clients. After the release of FedGEN, some other recent work focuses on improving the generator's performance. For instance, \textcite{zhang2022fine} first generates pseudo-data to train the generator and then utilizes the hard samples to train the global model simultaneously. Moreover, customized label sampling and label-wise ensemble algorithms are designed to boost the convergence of the distillation process. In addition to the above methods that utilize knowledge from the model’s outputs, requiring a distillation process at the server, using the feature representations is an emerging approach. \textcite{tan2022fedproto} present FedProto, which aggregates the abstract class prototypes collected from clients and then sends the global prototypes back to all clients to regularize the training of local models. The training on each client aims to minimize the classification error on local data while keeping the resulting local prototypes sufficiently close to the corresponding global ones. After that, \textcite{tan2022federated} design FedPCL to conduct contrastive learning during local training, allowing clients to share more class-relevant knowledge from local and global prototypes. While outperforming FedProto, FedPCL induces larger communication overhead due to the inter-client sharing of local prototypes. 

\paragraph{\textbf{Discussion}}
In this subsection, we have discussed state-of-the-art solutions for reducing the burdens on the network. \textit{Quantization}, \textit{Sparsification}, \textit{Factorization}, and \textit{Federated Distillation} are four primary approaches that have been well explored in the literature. However, each of them has its limitations. In general, all four approaches sacrifice the model capacity to achieve efficiency in communication, thus degrading the model performance. More specifically, \textit{Quantization} methods rely on advanced quantizers, therefore creating additional memory footprint and computational bottleneck at clients, \textit{Sparsification} and \textit{Factorization} methods could be overly complicated to be applied for advanced ML models. In contrast, \textit{Federated Distillation} methods typically require the server to perform more computation rather than only aggregating the global model. Remarkably, the emerging approach of sharing knowledge through abstract class prototypes rather than the model’s parameters or the model’s outputs has a large potential to be developed further since it faces no such issues. These discussed problems open numerous challenges and opportunities for future research in reducing the burdens on the network. 

\section{Communication-Efficient FL Architectures}
In prior sections, centralized approaches for FL had been discussed. However, the reliance on a central node poses potential problems such as latency, risk for a single point of system-wide failure, and trustworthiness concerns of the central FL server \cite{martinez2023decentralized}. Decentralized federated learning (DFL) was introduced in 2018 by He et al. \cite{he2018cola}, which involves decentralized model aggregation between neighboring nodes, which minimizes reliance on centralized architectures. This section will discuss two main DFL architectures categories: \textit{Hierachical FL} and \textit{Peer-to-Peer FL}.

\begin{figure}[!t]
\centering
\includegraphics[keepaspectratio, width=\textwidth]{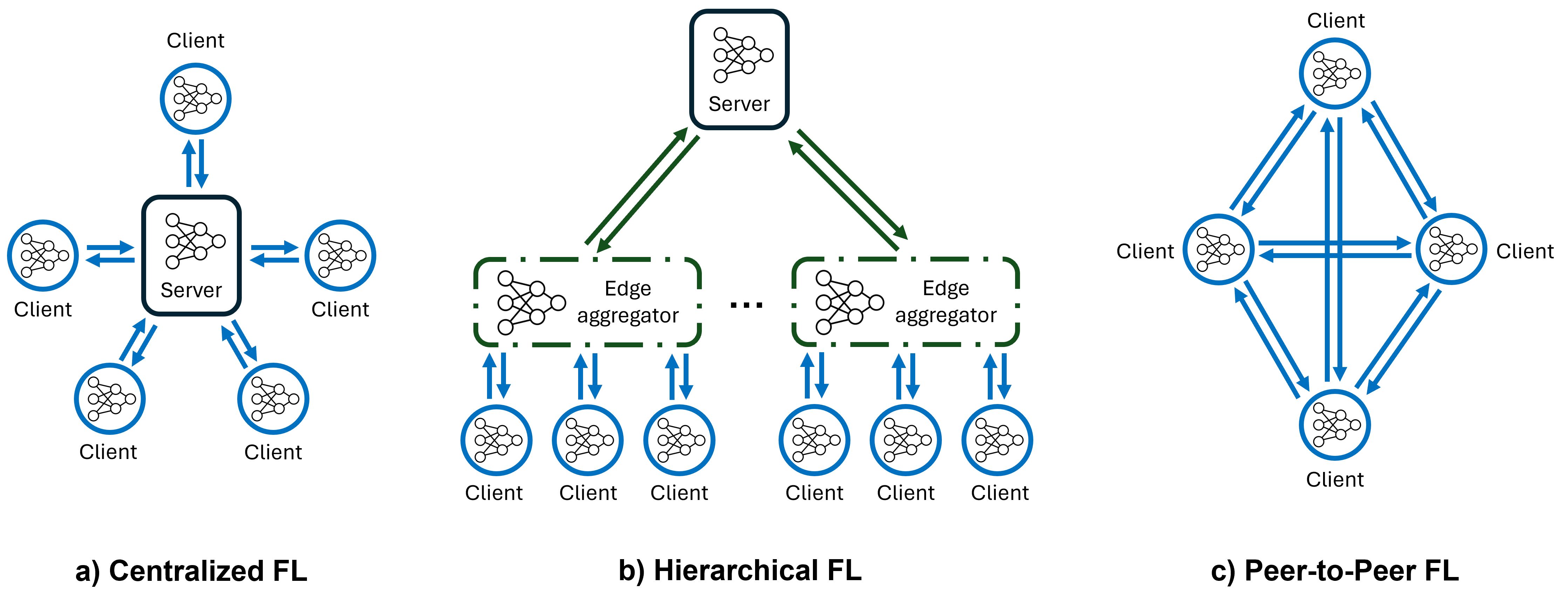}
\caption{Overview of different FL architectures. (a) The common centralized FL architecture is where each client communicates directly with a central cloud server. (b) A standard hierarchical FL architecture where clients communicate with an edge server for aggregation instead, and the edge servers will communicate with the central cloud server for global aggregation. (c) A standard Peer-to-Peer FL architecture is where clients communicate with each other, often without the need for a local or global aggregator. }
\label{fig-types}
\end{figure}

\subsection{Hierarchical FL}
Liu et al. proposed an edge-cloud hierarchical framework called HierFAVG \cite{liu2020client} to tackle communication resource bottlenecks in FL. HierFAVG has a three-layer architecture - clients, edge servers, and a cloud server. The edge servers aggregate models from their local clients, and the cloud server then aggregates the models from the edge servers. Clients perform local updates on their models based on their local data. After every $k_1$ local updates on each client, the client refines its local model parameters through edge aggregation where each edge server aggregates the models from its connected clients. This partial aggregation at the edge level helps consolidate information from multiple clients within the same edge server. After every $k_2$ edge aggregation, the cloud server aggregates the models from all edge servers. The cloud-level aggregation combines the partially aggregated models from different edge servers, enabling a higher-level information aggregation. The method offers an advantage in communication over centralized FL via less expensive communications to an edge server while also performing better than distributed learning through model aggregations. 

\subsection{Peer-to-Peer FL}
Compared to Hierarchical FL, clients in Peer-to-Peer FL communicate directly.
DFedAvgM \cite{sun2023decentralized} extends the centralized FedAvg algorithm to a decentralized setting, where clients are connected by an undirected graph and communicate with their neighbors instead of a central or edge server. In DFedAvgM, each client performs multiple local SGD with momentum iterations and then sends the updated parameters to its neighbors, which is determined by the communication graph associated with a mixing matrix. The clients then update their local parameters by taking a weighted average of the received parameters from neighbors. To further reduce the communication cost between clients, the paper also proposes a quantized version of DFedAvgM, where each client sends quantized parameter updates to its neighbors, which reduces the amount of information that needs to be communicated between clients. By quantizing the model updates, the amount of data transmitted between clients is reduced, which is particularly beneficial when the number of neighbors a client communicates with grows. This efficient communication via quantization helps alleviate the bottleneck caused by client-client communications, improving the overall efficiency of the algorithm. In essence, the quantized DFedAvgM optimizes communication in distributed training by transmitting quantized model updates between clients, thereby reducing the amount of data exchanged and improving the overall efficiency of the decentralized FL process.

HL-SGD \cite{guo2022hybrid} leverages peer-to-peer and device-to-server communications to accelerate learning. Devices are grouped into disjoint clusters that can communicate efficiently with each other. The first option is natural clustering based on device location or network topology in which devices that are physically close to each other or belong to the same local-area network (LAN) domain can be grouped into the same cluster, as they will have high-bandwidth peer-to-peer connectivity. The second is clustering based on geographic locations, where mobile devices can be grouped into clusters based on their geographic locations, such that devices in the same cluster are close to each other and can communicate efficiently. We can also consider random partitioning if there is no natural way to group devices, as long as the devices within each cluster have good peer-to-peer connectivity, allowing them to exchange information during the local update efficiently and averaging steps of the HL-SGD algorithm. In each round, devices first perform local SGD updates within their cluster. Then, a subset of devices are sampled from each cluster to upload their models to the central server. The key idea is to leverage the fast peer-to-peer communication within clusters to perform efficient local model updates while only selectively sending a subset of the local models to the server to reduce the overall communication cost compared to standard FL approaches. This hybrid model aggregation scheme, which uses both inexpensive peer-to-peer communication and infrequent centralized aggregation, aims to strike a balance between model accuracy and training time. 

\label{sec-decentralized}

\begin{figure}[!t]
\centering
\includegraphics[keepaspectratio, width=0.6\textwidth]{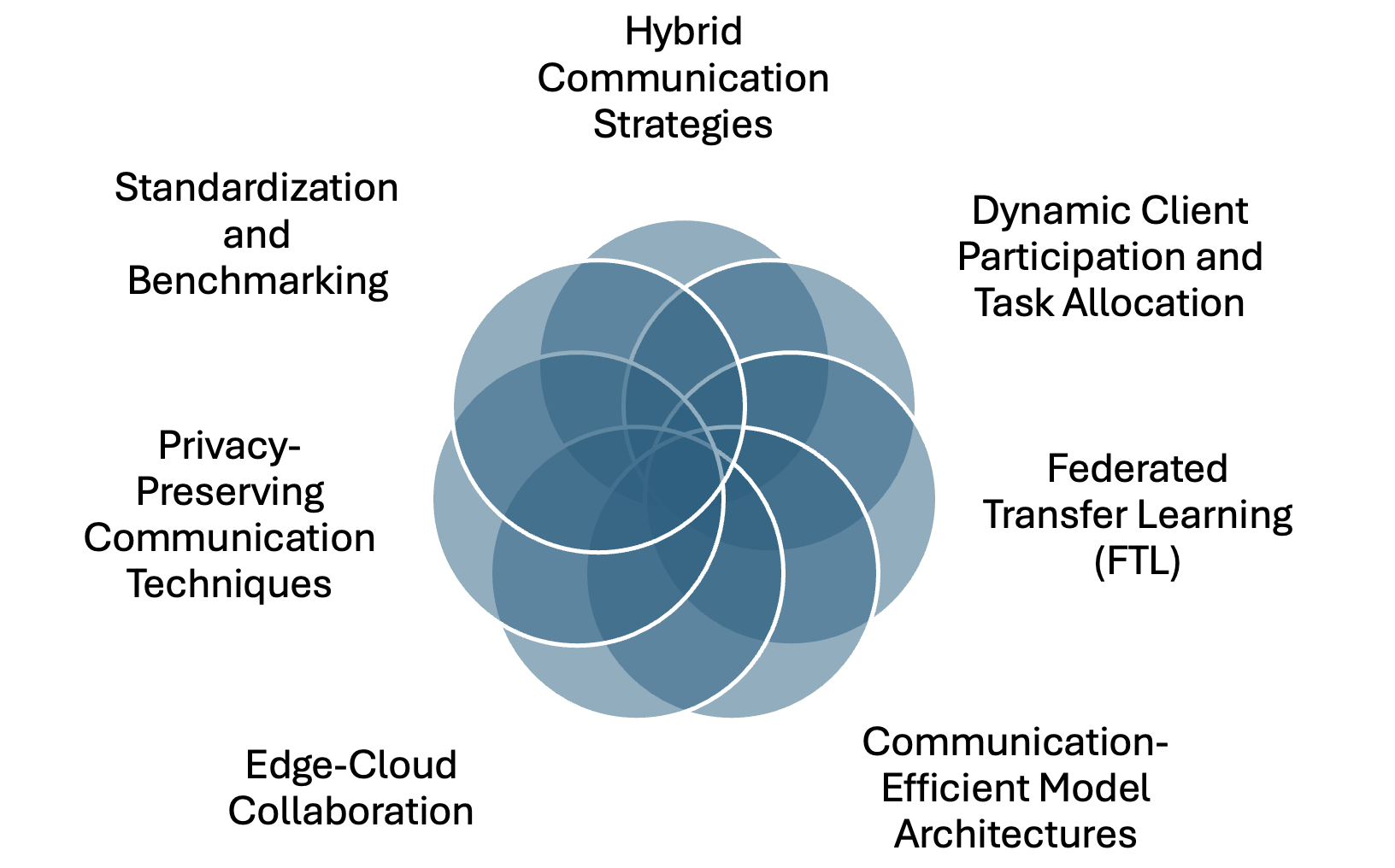}
\caption{Overview of Future Directions in Practicality of Federated Learning.}
\label{fig-direction}
\end{figure}

\section{Future Directions and Discussions}

FL systems face a significant challenge due to the variability in network conditions and heterogeneous device capabilities, leading to inefficiencies in communication. The network environments in which devices operate can vary drastically, impacting the speed and reliability of data transmission. Factors such as bandwidth constraints, latency, packet loss, and network congestion can contribute to communication inefficiencies in FL settings. Moreover, the devices participating in FL may have diverse hardware specifications, including differences in processing power, memory capacity, and battery life. These variations can affect the efficiency of communication protocols and the ability to handle large model updates effectively. This section will present the potential future directions concerning research into FL-communication efficiency from a practical perspective and discussions related to the impact of communication efficiency on real-world FL deployment.

\subsection{Future Directions}
In this survey, we have identified several promising future directions (as shown in Fig.\ref{fig-direction}) that can further advance the goal of improving communication efficiency in FL while addressing the existing challenges for real-world development.

\begin{itemize}
    \item \textbf{Hybrid Communication Strategies}: Explore hybrid communication strategies that combine the benefits of centralized and decentralized approaches. For example, hierarchical FL architectures \cite{liu2020client,briggs2020federated} could be developed where local updates are first aggregated within smaller clusters of devices before being sent to the central server. This reduces the network's burden while allowing for global model updates. However, managing hierarchical FL architectures requires sophisticated coordination mechanisms to ensure local updates from smaller clusters are correctly aggregated and synchronized with the central server \cite{yang2021h}. This added complexity can introduce delays and potential bottlenecks. Furthermore, the multiple layers of aggregation (local clusters to a central server) can increase latency. Hence, this direction requires a careful balance of design choices, robust coordination protocols, and advanced algorithms to ensure efficient and reliable FL implementations using hybrid communication strategies.

    \item \textbf{Dynamic Client Participation and Task Allocation}: Develop algorithms and frameworks for dynamic client participation and task allocation based on factors such as device capabilities, data relevance, and network conditions. This adaptive approach can optimize communication efficiency by dynamically adjusting the set of participating clients and the tasks assigned to them in each communication round \cite{zhang2021adaptive}. These algorithms and frameworks must dynamically evaluate and rank clients, which can be computationally intensive and require real-time data.

    \item \textbf{Federated Transfer Learning (FTL)}: Extend FL to support transfer learning paradigms, where knowledge learned from one task or domain is transferred to another related task or domain \cite{saha2021federated}. This reduces the need for extensive communication by leveraging pre-trained models or knowledge transferred from other devices, especially in scenarios with limited data availability or communication resources. Pre-trained models may have different neural network architectures, making it challenging to combine them directly. Techniques like model surgery, which involves careful surgical operations on the model architectures, may be required to ensure compatibility and enable knowledge transfer between heterogeneous architectures.

    \item\textbf{Communication-Efficient Model Architectures}: Design communication-efficient model architectures specifically tailored for FL settings \cite{nguyen2022toward}. This includes developing lightweight model architectures, parameter-sharing techniques, and model distillation methods that reduce the size of updates transmitted over the network while maintaining model performance. Reducing the size of model updates while maintaining or improving model performance is a significant challenge. Lightweight architectures must be carefully designed to ensure they do not sacrifice the accuracy or robustness of the model.

    \item \textbf{Edge-Cloud Collaboration}: Explore collaborative FL frameworks that effectively leverage edge and cloud resources. Edge devices can perform local model updates and collaborate with cloud servers for global model aggregation, minimizing communication overhead and latency. This requires efficient synchronization and coordination mechanisms between edge and cloud components.

    \item \textbf{Privacy-Preserving Communication Techniques}: Research novel privacy-preserving communication techniques, such as secure multiparty computation (SMC) \cite{goldreich1998secure} and homomorphic encryption \cite{yi2014homomorphic}, to protect data privacy during model aggregation and update transmission. These techniques enable FL systems to securely aggregate updates without exposing raw data, enhancing communication efficiency while preserving privacy. However, both SMC and homomorphic encryption are computationally intensive. Hence, research can focus on optimizing homomorphic encryption schemes to reduce computational and communication overhead. This could involve developing more efficient encryption algorithms or implementing partial homomorphic encryption tailored for specific operations used in FL.

    \item \textbf{Standardization and Benchmarking}: Different applications may prioritize different aspects of communication efficiency, such as latency, bandwidth usage, or energy consumption. One promising direction is to establish standardized benchmarks and evaluation metrics for assessing the communication efficiency of FL algorithms across different scenarios and applications. This facilitates fair comparison and benchmarking of existing methods and encourages the development of new techniques that improve communication efficiency.

\end{itemize}

\subsection{Discussions}
The impact of communication efficiency on real-world FL deployment is significant and multifaceted. It influences the FL system's performance, scalability, and practical feasibility. For instance, communication efficiency directly affects the latency of model updates transmitted between the central server and participating devices. Improvements in communication efficiency lead to reduced latency, enabling faster model convergence and more responsive FL systems. Also, efficient communication strategies minimize the data transmitted over the network during FL training rounds. Reducing bandwidth requirements is crucial for accommodating large-scale FL deployments with limited network bandwidth or IoT-edge devices with constrained data plans. 

Communication efficiency is essential for scaling FL systems to many participating devices or clients. Efficient communication protocols and algorithms ensure the system can handle increasing numbers of devices without overwhelming the network or the central server. In resource-constrained environments such as mobile devices or edge computing platforms, efficient communication reduces energy consumption during FL training. Energy-efficient FL deployments can prolong device battery life and reduce operational costs by minimizing the amount of data transmitted and the frequency of communication rounds. FL systems deployed in real-world settings must contend with variable network conditions, including bandwidth, latency, and reliability fluctuations. Communication efficiency measures, such as adaptive scheduling and error-handling mechanisms, help maintain system performance and reliability under diverse network conditions. 

Efficient communication protocols are crucial in preserving data privacy and security in FL deployments. By minimizing the transmission of sensitive data and leveraging privacy-preserving techniques such as federated averaging or secure aggregation, communication efficiency helps protect user privacy while enabling collaborative model training. Optimizing communication efficiency reduces the infrastructure and operational costs associated with FL deployments. By minimizing data transmission costs and network overhead, organizations can deploy FL systems cost-effectively and achieve better returns on investment.

Interdisciplinary collaboration among experts from diverse fields is necessary to address the multifaceted challenges of practical FL deployment from a communication perspective. Such collaboration can lead to the development of holistic solutions that integrate communication-efficient techniques with privacy-preserving mechanisms, distributed architectures, edge computing strategies, and domain-specific requirements. For instance, expertise in communication networks and protocols is essential for developing efficient, scalable, and reliable communication strategies for FL systems. Collaborating with experts in distributed systems, parallel computing, and decentralized architectures is crucial for designing scalable and fault-tolerant FL systems. ML experts are needed to develop effective model compression, quantization, and aggregation techniques tailored for communication efficiency. In addition, collaboration with domain experts from various application areas (e.g., healthcare, finance, IoT) is essential to understand the unique communication requirements, constraints, and privacy concerns in those domains.


\section{Conclusions}
FL represents a transformative approach in machine learning by facilitating collaborative training across multiple decentralized participants or devices without necessitating data centralization. This paradigm shift is particularly advantageous in sectors prioritizing data privacy. Despite its potential, the practical deployment of FL is significantly challenged by communication overhead, which is the main bottleneck that hinders its scalability and efficiency in real-world applications. Since there was no comprehensive survey on the practicality of federated learning (FL) in the existing literature, this survey provides an in-depth discussion on implementing FL in real-world applications, focusing on communication aspects. We begin by defining the cost of communication in terms of the total communication time, and the volume of data exchanged. We then examine strategies to enhance communication efficiency, such as reducing the number of communication rounds, minimizing the number of participating clients, and employing model compression techniques like quantization, sparsification, factorization, and federated distillation. Following this, we present communication-efficient federated learning (FL) methods and discuss various FL architectures designed to mitigate communication challenges. Finally, we highlight research challenges and propose future directions for practical deployment


In summary, addressing the communication bottleneck is essential for realizing the full potential of FL in practical settings. FL can be effectively scaled and deployed across various domains requiring privacy-preserving, decentralized machine learning by developing communication-efficient methods and exploring future directions for communication efficiency. This comprehensive survey underscores the importance of communication efficiency and suggests future research and development to overcome current limitations and enhance the practical applicability of FL.

\section*{Acknowledgements}

This work is funded by the Deutsche Forschungsgemeinschaft (DFG, German Research Foundation) - project number 453130567 (COSMO), by the Horizon Europe Research and Innovation Actions under grant number 101092908 (SmartEdge) and by WIDERA programme under the grant agreement No. 101079214 (AIoTwin); by the Federal Ministry for Education and Research, Germany under grant number 01IS18037A (BIFOLD).


\clearpage
\printbibliography
\end{document}